%% file: main.tex
\documentclass[10pt,twocolumn,letterpaper]{article}
\usepackage[pagenumbers]{cvpr}
\input{bench_preamble}
\definecolor{cvprblue}{rgb}{0.21,0.49,0.74}
\usepackage[pagebackref,breaklinks,colorlinks,allcolors=cvprblue]{hyperref}
\usepackage{flushend}

\usepackage[most]{tcolorbox}
\colorlet{promptaccent}{cvprblue!85!black}
\colorlet{promptback}{cvprblue!5!white}
\newtcolorbox{promptbox}[3][]{%
  enhanced,
  colback=promptback, colframe=promptaccent, colbacktitle=promptaccent, coltitle=white,
  boxrule=0.4pt, arc=2pt, outer arc=2pt,
  left=6pt, right=6pt, top=5pt, bottom=5pt, toptitle=2pt, bottomtitle=2pt, lefttitle=6pt, righttitle=6pt,
  fonttitle=\sffamily\bfseries\scriptsize,
  title={#2\hfill{\mdseries\itshape #3}},
  fontupper=\footnotesize\sffamily,
  before upper={\raggedright\setlength{\parskip}{3pt}\setlength{\leftmargini}{1.1em}},
  before skip=6pt, after skip=8pt,
  #1}
\newcommand{\ptool}[1]{{\ttfamily\color{cvprblue!55!black}#1}}
\newcommand{\tool}[1]{\texttt{#1}}
\makeatletter\let\inputtab\@@input\makeatother

\def\confName{CVPR}
\def\confYear{2027}
\title{Training Vision Models for LLM-Driven Weight-Space Programming}
\author{Timo Ropinski\\
Ulm University\\
{\tt\small timo.ropinski@uni-ulm.de}
\and
Sebastian Kranz\\
Ulm University\\
{\tt\small sebastian.kranz@uni-ulm.de}}
\begin{document}
\maketitle

\begin{abstract}
It has been demonstrated that, in certain settings, large language models can perform weight-space programming to directly modify the weights of neural networks. However, it is unknown how dedicated training of these networks can contribute to the successes of weight-space programming. We present a benchmark of LLM-driven weight-space programming on vision models, in which five LLMs edit convolutional networks and vision transformers on relabeling and error-correction tasks on Waterbirds, CIFAR-100, PACS and Entity-30. We report more than 2{,}600 such edit processes, whereby each edit is scored on held-out images for success, side effects and accuracy loss. We compare standard training with label smoothing, gradient-based editable training, and our Edit-Ready Training (ERT), a novel method that rehearses the direct, closed-form weight edits LLMs perform. We show that preparing networks for editing raises the success of LLM edits substantially where standard networks are hard to edit. ERT further achieves the highest mean score for GPT-6 Astra and Claude Opus 5.5 across all evaluated baselines. We will release our LLM testing framework alongside all tasks, editing trajectories and scores. Project page: \url{https://viscom.uni-ulm.de/wsp}
\end{abstract}

\begin{figure*}[t]
  \centering
  \includegraphics[width=\textwidth]{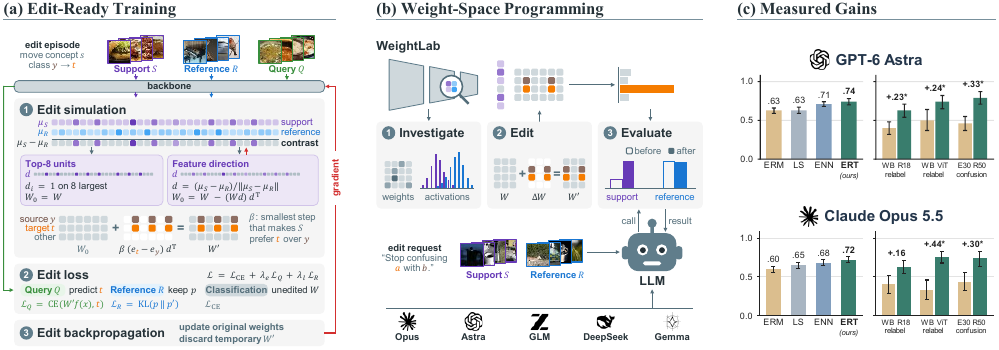}
  \caption{\textbf{Training vision models for weight-space programming.}
  \textbf{(a)} Edit-Ready Training (ERT) simulates edits during training to make vision models better editable by LLMs.
  \textbf{(b)} Through our WeightLab an LLM agent can iteratively edit a vision model's weights to change its behavior.
  \textbf{(c)} ERT improves vision model editability through LLMs, which is indicated by an increased mean net edit score over all nine settings, and the gain of ERT over ERM on the three settings that are hardest (bars: 95\% bootstrap confidence; $^*$: $p<0.05$).}
  \label{fig:ert-overview}
\end{figure*}

\section{Introduction}
\label{sec:intro}

Deployed image classifiers often need small, specific changes: a fine-grained concept has to move to another class, a recurring confusion between two classes has to be repaired, or a rare subgroup is misclassified. Collecting data and fine-tuning is costly and can change behavior far beyond the requested fix. Model editing methods instead compute a targeted weight change for one request~\cite{santurkar2021editing,meng2022locating,mitchell2022fast}. Recently, general-purpose large language models (LLMs) have begun to edit networks directly, a practice called \emph{weight-space programming}~\cite{takehi2026sparks}. Here, an LLM agent inspects activations and weights, finds the units that respond to a concept, writes a targeted change and checks its effect with forward passes, all without gradient descent.

The networks such agents edit are, however, trained for accuracy alone. Nothing in their training asks that a concept be represented in a form that a short, targeted edit can isolate. It is therefore unknown how well LLM agents edit standard vision models, and whether training can make networks easier for them to edit.

We study both questions for image classification (see \cref{fig:ert-overview}). Blinded LLM agents receive an edit request with a few support images, edit any weight of the network through a tool interface, without gradients, under a fixed budget of forward passes, weight reads and weight edits, and are scored on held-out test images for success and side effects. During these edit processes, they relabel a fine-grained concept, fix a confusion between two classes, or fix a misclassified subgroup, on Waterbirds~\cite{sagawa2020group}, CIFAR-100~\cite{krizhevsky2009learning}, PACS~\cite{li2017deeper} and Entity-30~\cite{santurkar2021breeds}, with ResNet-18, ResNet-50~\cite{he2016deep} and ViT-B/16~\cite{dosovitskiy2021image} classifiers. Five LLMs act as editors, led by GPT-6 Astra and Claude Opus 5.5, to edit networks that are trained in four ways: standard empirical risk minimization (ERM), label smoothing (LS), gradient-based editable training (ENN)~\cite{sinitsin2020editable}, and Edit-Ready Training (ERT), which we introduce. Our pre-study revealed that agents mostly change the classifier head along the contrast between support and other images, and ERT rehearses such edits during training, so that the features become easy to edit this way.

On four of the nine tested settings formed by dataset, task and backbone, standard networks are already easy to edit, and dedicated training adds little on average. Where they are hard to edit, ERT helps substantially for both main editors. On the Entity-30 confusion fix it raises the net score from 0.46 to 0.79 for Astra and from 0.44 to 0.74 for Opus, and on Waterbirds with ViT-B/16 from 0.50 to 0.74 and from 0.33 to 0.76. While none of the compared training methods is best across the board, ERT has the highest mean net score for both leading editors. Thus, our main contributions are as follows:
\begin{itemize}\setlength\itemsep{1pt}
\item We present a systematic study of how well LLM agents edit trained vision classifiers, and how the classifier training affects edit quality.
\item We propose Edit-Ready Training (ERT), which delivers the highest mean net edit score for GPT-6 Astra and Opus 5.5 among all tested baselines.
\item We release a vision weight-space programming benchmark, alongside more than 75{,}000 logged tool calls describing scored edits from five LLM editors.
\end{itemize}

\section{Related Work}
\label{sec:related}

\noindent\textbf{Weight-space analysis.}
Weight-space analysis treats the weights of trained networks as data. Test accuracy or training hyperparameters can be predicted from the weights alone~\cite{eilertsen2020classifying,unterthiner2020predicting}, representations learned on model zoos capture model characteristics without labels~\cite{schurholt2021self,schurholt2022model}, architectures for learning on weights respect the permutation symmetries of neurons~\cite{navon2023equivariant,zhou2023permutation}, and hypernetworks generate weights~\cite{ha2017hypernetworks}. Interpretability research can assist the analysis of networks by assigning visual concepts to units~\cite{bau2017network}, combining weights between units into circuits~\cite{olah2020zoom}, and by describing units in natural language~\cite{hernandez2022natural,oikarinen2023clip}.

Language models have recently taken over parts of this analysis. Agents with interpretability tools run experiments on vision models and explain their units~\cite{shaham2024maia}, and language models have been trained to read weights~\cite{kabra2026weightoracles}. Frontier LLMs identify small networks from their raw weights without such training~\cite{takehi2026sparks}, although their readings of small image networks break under label permutation~\cite{ahmed2026memorization}.

\noindent\textbf{Neural network editing.} Network editing changes the behavior of a trained network on a specific input or concept while leaving the rest intact. Most methods compute the weight change with an algorithm: fine-tuning, learned hypernetworks that turn a gradient into an edit~\cite{decao2021editing,mitchell2022fast}, or closed-form low-rank updates at located layers~\cite{meng2022locating,meng2023memit}. In vision, prediction rules are rewritten at a chosen layer of a classifier~\cite{santurkar2021editing} or a generative model~\cite{bau2020rewriting}, and concepts are erased from diffusion models~\cite{gandikota2023erasing}. Task arithmetic adds and subtracts weight differences~\cite{ilharco2023editing}, linear projections remove a concept direction from the features~\cite{ravfogel2020null}, and retraining only the last layer corrects reliance on spurious features~\cite{kirichenko2023last}, which makes the classifier a natural place for small edits.

Weight-space programming~\cite{takehi2026sparks} replaces the editing algorithm with an LLM that decides itself what to inspect and what to write. This offloads editing to LLM agents that carry out technical work autonomously~\cite{huang2024mlagentbench}. Existing work~\cite{takehi2026sparks} judges an edit only by its effect on the target behavior. Our benchmark also scores side effects and accuracy cost on held-out images, blinds the agent to the origin of the network, and compares several agents under one protocol.

Few works target editability during training. Editable Neural Networks (ENN)~\cite{sinitsin2020editable} meta-train a network, in the spirit of MAML~\cite{finn2017model}, so that a few gradient steps on new examples correct its predictions without harming others. We keep this structure but rehearse the edits that LLM agents perform. Apart from dedicated techniques, ordinary training choices also affect editability~\cite{liu2025dropdropout}. Label smoothing~\cite{szegedy2016rethinking,muller2019label} produces less extreme logits, and sparse final layers make a classifier easier to inspect and debug~\cite{wong2021leveraging}. To our knowledge, none of these methods has been evaluated with LLM editors.

\section{Method}
\label{sec:method}

Our approach has two parts, as shown in \cref{fig:ert-overview}. First, Edit-Ready Training (ERT) prepares a network for later edits, by simulating head weight edits, computing the errors of the edited network, and backpropagating them through the edit into all weights. Second, Iterative Weight-Space Programming (WSP) edits the trained network afterwards in an iterative process. An LLM receives an edit request and a few images, and uses our WeightLab sandbox server to inspect the network, modify its native weights and check the result. We describe WSP first, because the rules of its editing protocol determine which edits ERT performs during training.

\subsection{Iterative Weight-Space Programming}
\label{sec:interface}

To systematically investigate weight-space programming of vision models, an LLM edits a trained network through WeightLab, our sandbox environment that exposes the network and enforces a fixed editing protocol (\cref{fig:ert-overview}, center). The protocol serves five purposes. It keeps the edits direct, such that the agent has no gradients, cannot train the network and cannot add layers or adapters, so it cannot fall back on fine-tuning, which would make it a gradient-based editor. It further makes editors and training methods comparable, as every editor uses the same tools, budget and system prompt (\cref{sec:supp-protocol}). Blinding keeps an editor from recognizing how a network was trained. We record every edit by logging all calls and changes, which makes the edits analyzable (see \cref{sec:editpatterns}). Finally, a budget limits trial and error on the few support images, which would otherwise lead to overfitting.

For our setup, we consider a classifier with a backbone that computes features $f(x)\in\mathbb R^d$ and a linear head with weights $W$ and bias $b$, so that the logits are $z(x)=Wf(x)+b$. An edit request asks to move a concept $s$, currently assigned to class $y$, to a target class $t$, while all other predictions stay as they are. More generally, an edit request consists of a few \emph{support} images labeled with the class they should receive and a set of \emph{reference} images labeled with the network's current predictions, which should not change. Besides relabeling a concept, we consider two kinds of requests that correct existing errors: fixing a confusion, where images of one class are predicted as another class, and fixing a subgroup, where images of one class from a particular domain, such as sketches, are misclassified (\cref{fig:tasks}).

Each edit process addresses a single edit request, such as ``move this concept to class $t$'', and starts with a fresh agent. Each request provides a few support images labeled with the desired class (five for relabeling, eight for the fix tasks) and 64 reference images, labeled with the network's original predictions to indicate what should remain unchanged. The agent never sees the images themselves, and it observes only how the network responds to them.

Within each edit process, the agent can iteratively inspect, modify, and evaluate the network. It inspects predictions, layer activations, feature contributions to logits, and parameter values. It then modifies weights by setting, adding to, or scaling individual entries, rows, channels, or whole tensors, anywhere in the classification head or the backbone. The protocol does not restrict which parameters are changed or how, the only limits are the number of calls and the size of a single edit call. During the process, the agent can probe the modified network on the provided images, revise its changes, undo the last edit, or reset the weights.

The agent submits these tool calls through model responses, which can contain several of these calls. Each edit process is limited to 20 forward probes, 30 weight reads, and 8 edit calls. A single edit call may change up to 65{,}536 parameters, and a single read returns at most 512 values. This write parameter budget is enough to rewrite the entire head of every network in one call (the largest, ResNet-50 on Entity-30, has 61{,}470 parameters) and any row, channel, bias or normalization layer of the backbone. At the same time, the read parameter budget, stops the agent from exfiltrating and retraining the network. Undo and reset do not count towards the edit budget, and do not restore it either. During editing, the agent further sees only an anonymous request ID, never receives images from the test set, and has no information about how the network was trained. Its system prompt encourages targeted edits that generalize beyond the provided examples (full text in \cref{sec:supp-objectives}). An edit process ends when the agent submits its modified weights. If the agent stops without submitting or reaches the limit of 150 model responses, the original weights are evaluated.

\subsection{Edit-Ready Training}
\label{sec:ert}

Standard training minimizes the cross-entropy $\mathcal L_{\rm CE}$ on the training labels and does not enforce editability. Edit-Ready Training (ERT) in contrast rehearses during training the edits that LLM agents make. To enable an agent to edit any parameter of the head or the backbone, our editing protocol does not constrain these edits during weight-space programming. However, during training the rehearsed edits are meant to simulate real Opus edits. Therefore, during a small pre-study, we have tasked Opus to edit ERM-trained networks. We found that Opus mainly (>90\%) changes head weights, an observation which we could also confirm later during our full edit analysis (see \cref{sec:editpatterns}). Furthermore, the performed edits were mostly in a rank-one manner, i.e., confined to the rows of the source and the target class, and concentrated on the few features that differ most between the support and the reference images. To simulate such edits during training, each training step draws a few \emph{edit episodes} that simulate an edit request to move a concept $s$ of class $y$ to a target class $t$. During these episodes, four \emph{support} images of $s$ define the edit, four unseen \emph{query} images of $s$ check that it generalizes, and \emph{reference} images of other concepts from the batch check that their predictions stay unchanged. Edit episodes draw $s$ only from training concepts, concepts that editors later changed during testing are never used. During the training process, all edit episodes pass through the three ERT steps (see \cref{fig:ert-overview} (a)), which we describe below.

\paragraph{Edit simulation.} For each episode, ERT computes a temporary head $W'$ that contains a simulated edit of the current head $W$. The edit is built from the mean backbone features, i.e., the inputs of the head, of the support images $S$ and the reference images $R$,
\begin{equation}
\mu_S=\frac{1}{|S|}\sum_{x\in S} f(x), \qquad \mu_R=\frac{1}{|R|}\sum_{x\in R} f(x).
\label{eq:means}
\end{equation}
Their contrast $\mu_S-\mu_R$ is large in the feature units that respond to the support images but not to the reference images. From this contrast, ERT derives an edit vector $d$ in feature space, by one of the two pre-study-identified rules, given below, and writes it into the rows of the source class $y$ and the target class $t$:
\begin{equation}
W' = W_0 + \beta\,(e_t-e_y)\,d^\top ,
\label{eq:ert}
\end{equation}
where $W_0$ is the head the edit starts from (defined per rule below), $\beta\ge 0$ is the step size, and $e_t$, $e_y$ are the one-hot vectors that select the rows of $t$ and $y$. Relative to $W_0$, the edit adds $\beta d$ to the target row and subtracts it from the source row, and leaves all other rows unchanged. For an image $x$, the logit of $t$ therefore changes by $\beta\,d^\top f(x)$ and the logit of $y$ by $-\beta\,d^\top f(x)$. Images whose features align with $d$ ($d^\top f(x)>0$), as the support images typically do, gain evidence for $t$ and lose it for $y$, while images with features orthogonal to $d$ keep all their logits. Based on our pre-study analysis of Opus edits, we have identified closed-form representations of the two most frequently used edit rules:
\begin{itemize}\setlength\itemsep{1pt}
\item \emph{Top-8 units:} $d$ is 1 on the $k=8$ feature units with the largest entries of $\mu_S-\mu_R$ and 0 elsewhere, and $W_0=W$, giving a sparse edit of 16 numbers.
\item \emph{Feature direction:} $d=(\mu_S-\mu_R)/\lVert\mu_S-\mu_R\rVert$, and $W_0=W-(Wd)d^\top$ first removes what every class currently reads along this direction, before the edit sets the readout along $d$ to $\beta$ for $t$, $-\beta$ for $y$ and zero for all other classes. The total change to $W$ is a dense rank-one edit that shifts every row by a multiple of $d$.
\end{itemize}
The step size $\beta$ is the smallest one that makes the support images on average prefer $t$ over $y$ by a margin of $\delta=1$,
\begin{equation}
\beta=\mathrm{clip}_{[0,10]}\!\left(\frac{m_S+\delta}{2\max(\mu_S^\top d,\,0.1)}\right),
\end{equation}
where $m_S$ is the mean logit margin of $y$ over $t$ on the support images under $W_0$. The edit shifts this margin by $2\beta\,\mu_S^\top d$ in favor of $t$, which gives the fraction. The floor of $0.1$ and the clipping to $[0,10]$ keep $\beta$ bounded when the support images barely align with $d$. This mirrors an agent that scales its edit until the probes show the support images flipped.

\paragraph{Edit loss.} The edited network is judged on two sets of images. The \emph{query} loss asks that the query images, held-out images of the concept that the edit has not seen, are now predicted as the target class, and the \emph{reference} loss asks that the predictions on the reference images match those of the original network:

\begin{equation}
\begin{aligned}
\mathcal L &= \mathcal L_{\rm CE}
+\lambda_e\,\mathbb E_{Q}\,\mathrm{CE}\bigl(W'f(x)+b,\,t\bigr)\\
&\quad+\lambda_l\,\mathbb E_{R}\,\mathrm{KL}\bigl(p(x)\,\Vert\,p'(x)\bigr),
\end{aligned}
\label{eq:loss}
\end{equation}
where $\mathcal L_{\rm CE}$ is the standard cross-entropy on the training batch, and $p$ and $p'$ are the predictions before and after the simulated edit, with $p$ treated as a constant.%

\paragraph{Edit backpropagation.} The simulated edits are a fixed, closed-form function of the current network, so both post-edit losses are differentiable with respect to all weights, including the backbone. The selection of the top-8 units is held fixed in the backward pass, while $\beta$ and the feature direction pass gradients. The update is applied to the original weights and the temporary $W'$ is discarded, so no edit is ever kept. Training therefore does not make one particular edit work. Instead, it is meant to shape the features so that edits also work for concepts and images not seen during training, with the goal that each concept is kept in a few distinctive units or a clean direction. The extra cost per training step are the forward and backward passes over the episode images, which are appended to the training batch, and one closed-form edit per episode.

\section{Experimental Setup}
\label{sec:experiments}

\subsection{Network Training}
\label{sec:networktraining}

\begin{figure*}[t]
\centering
\includegraphics[width=\textwidth]{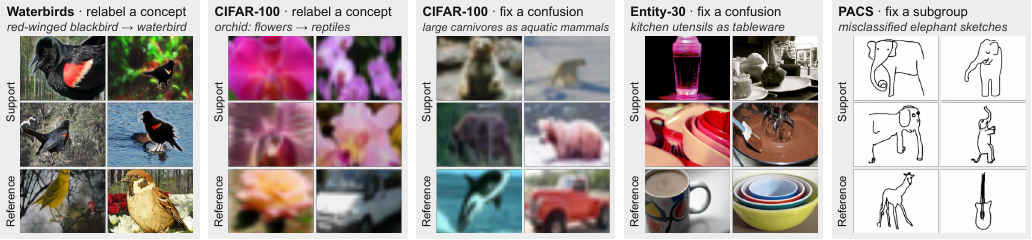}
\vspace{-12pt}
\caption{\textbf{Two image types are used during editing.} \emph{Support} images carry the class they should receive after the edit (concept to relabel, errors to correct). \emph{Reference} images are samples whose predictions must stay as they are.}
\label{fig:tasks}
\vspace{-4pt}
\end{figure*}

\paragraph{Vision model training.} All networks to be edited are initialized with ImageNet weights and are classification-fine-tuned on the dataset's classes: ResNet-18~\cite{he2016deep} and ViT-B/16~\cite{dosovitskiy2021image} on Waterbirds (100 epochs), CIFAR-100 (30 epochs) and PACS (50 epochs), and ResNet-50 on Entity-30 (20 epochs).

Since training compute and frontier model requests are limited, we train a single network per training method and setting, with the exception of multiple trained networks for the seed ablation in \cref{sec:ablations}. The intervals we report therefore reflect variation across edit processes, conditional on the trained networks. The seed ablation in \cref{sec:ablations} supports this design: the variation between networks trained with different seeds is negligible compared to the variation between edit processes on the same network.

\paragraph{Baselines.} We compare our ERT against three baselines. \emph{Standard empirical risk minimization (ERM)} minimizes cross-entropy only. \emph{Label smoothing (LS)} replaces the one-hot target with $0.8$ on the true class and $0.2$ spread over all classes~\cite{szegedy2016rethinking}, to obtain less peaked logits~\cite{muller2019label} such that a small change can move a decision. \emph{ENN} denotes Editable Neural Networks~\cite{sinitsin2020editable}. ENN is analogous to ERT, but simulates the edit with one or five gradient steps on the support images, $W'=W-\eta\nabla_W \mathbb E_S\,\mathrm{CE}(Wf(x)+b,t)$. While the original ENN paper allows for arbitrary edit locations, we choose the head parameters, as these were mainly edited by LLM editors in our pre-study (see \cref{sec:ert}). Like ERT, ENN backpropagates the post-edit losses into all weights.

LS and ENN were the strongest alternatives to standard training in a pool of ten methods, which we have filtered using scripted edits resembling real LLM edits. Other techniques in the pool included sharpness-aware minimization~\cite{foret2021sharpness}, dropout~\cite{srivastava2014dropout}, auxiliary concept labels~\cite{koh2020concept} and a sparse final layer~\cite{wong2021leveraging} (details in \cref{sec:supp-screen}).

\paragraph{Hyperparameter tuning.} ERT, as well as all baseline methods, were hyperparameter tuned with the same protocol on two ResNet-18 settings: Waterbirds and CIFAR-100. To keep compute needs low and the number of frontier model requests manageable, we also used scripted edits for selection. Furthermore, the ViT models reuse the ResNet-18 tuned hyperparameters, and PACS and Entity-30 reuse the CIFAR-100 tuned hyperparameters, across all methods. Search spaces and selected values can be found in \cref{sec:supp-tuning}.

\begin{table*}[t!]
\centering
\small
\setlength{\tabcolsep}{2pt}
\renewcommand{\arraystretch}{1.05}
\renewcommand{\cistack}[2]{#1\,{\fontsize{5.5}{6.5}\selectfont\color{black!55}[#2]}}
\caption{\textbf{Main Results.}
Net score ($\uparrow$) with 95\% bootstrap confidence of Astra \& Opus edits. WB=Waterbirds, C100=CIFAR-100, E30=Entity-30; R18/R50=ResNet-18/50; LS=label smoothing. On CIFAR-100 and PACS, ENN and ERT include label smoothing. $\Delta$ rows show difference of ERT vs. ERM and ERT vs. best other baseline with $p<0.05$ (paired sign-flip test) shown in bold.}
\label{tab:main}
\input{tables_bench/table1.tex}
\end{table*}

\subsection{Weight-Space Programming}
\label{sec:weightspaceprogramming}

\paragraph{Editing tasks.} We evaluate on three kinds of edit requests (see \cref{fig:tasks}). \emph{Relabel a concept:} a fine-grained concept moves to another class. The concept is a bird species on Waterbirds~\cite{sagawa2020group} (two classes) or a fine class on CIFAR-100~\cite{krizhevsky2009learning} (20 superclasses). \emph{Fix a confusion} (CIFAR-100, and Entity-30~\cite{santurkar2021breeds}, a 30-class subset of ImageNet~\cite{deng2009imagenet}): superclass $A$ is wrongly predicted as $B$, which should be corrected. \emph{Fix a subgroup} (PACS~\cite{li2017deeper}): misclassified sketches of one class should be corrected.

\paragraph{LLM editors and protocol.} The two main editors are GPT-6 Astra (high reasoning effort) and Claude Opus 5.5 (default effort), each restricted to the WeightLab tools. Both cover all nine settings with all training methods. Three open-weight editors test how far the results carry over to weaker models: GLM-5.3-Flash, DeepSeek-V4.1-Flash and Gemma-4-26B (detailed numbers in \cref{sec:supp-open}).

Each editor handles eight edit requests per setting and training method, each with its own images, on one trained network, or split over two networks on PACS and three on the Entity-30 confusion fix (three requests each), which the seed ablation in \cref{sec:ablations} uses. We pair results of the same editor for different training methods and compare two training methods with an exact two-sided sign-flip test on the paired differences. The main grid contains 292 scored edit processes of Astra, 308 of Opus and 757 of the three open-weight editors. Including ablations and pilots, the five editors made more than 2{,}600 scored edit processes with more than 75{,}000 tool calls, and the median edit process takes about three minutes for Astra and Opus (\cref{sec:supp-scale}).

\paragraph{Metrics.} The reported scores are computed on held-out test images that no editor sees. For relabeling, \emph{success} is the fraction of test images of the concept predicted as $t$, and the \emph{side effect} is the fraction of test images of all other concepts whose prediction flips. For the two fix tasks, success is the fraction of the targeted test errors that the edit corrects, and the side effect is the fraction of originally correct test images outside the target group that become wrong. The \emph{net score} as our primary measure reports success minus side effect, as also reported by Sinitsin et al.~\cite{sinitsin2020editable}.

\section{Results}
\label{sec:results}

\subsection{Main Results}
\label{sec:main}

\paragraph{Baseline comparison.}
\cref{tab:main} shows the net scores of the two main editors (details on open-weight editors are found in \cref{sec:supp-open}). ERT has the highest mean net score for both editors (Astra: 0.74, against 0.71 for ENN, 0.63 for LS and 0.63 for ERM; Opus: 0.72, against 0.68, 0.65 and 0.60). ERT's gain over ERM is $+0.12$ for both ($p<0.001$ over the paired edit processes), and for Opus it is the only gain that stays significant when each of the nine settings counts as one observation ($p=0.047$; ENN $p=0.17$), as each method helps strongly on only a few settings. ERT is ahead of ENN by $+0.03$ (Astra) and $+0.04$ (Opus) on average, which is not significant ($p=0.10$ and $p=0.07$). It is also never far below ERM: its largest loss is $0.02$ for both editors, and its weakest setting with Opus still reaches 0.61, against 0.33 for ERM, 0.48 for LS and 0.52 for ENN.

Per setting, ERT is the best or second-best method for Astra everywhere, whereas for Opus each preparation method is best on at least two settings, LS and ENN notably on CIFAR-100 relabeling ($+0.13$, $p=0.02$, with ResNet-18; $+0.10$, $p=0.008$, with ViT). Preparation loses significantly against ERM on three settings (LS for Astra on the CIFAR-100 confusion fix with ResNet-18 and on PACS with ViT, $-0.10$ and $-0.11$; ENN for Opus on PACS with ViT, $-0.16$). Directly compared, ERT beats ENN on Waterbirds with ResNet-18 for both editors ($+0.16$, $p=0.02$, and $+0.10$, $p=0.04$), and ENN is nowhere significantly ahead of ERT.

\paragraph{Editor differences.}
Astra and Opus are about equally strong and rank the networks alike (36 cells: mean net 0.68 and 0.66, rank correlation $0.78$; their gains from preparation correlate at $0.77$), so the conclusions do not depend on which of the two is used.

Of the open-weight editors, DeepSeek-V4.1-Flash matches the main editors on many of the completed cells. GLM-5.3-Flash and Gemma-4-26B are much weaker (0.45 and 0.30, against 0.68 for Opus), rank the networks unlike Opus ($0.02$ and $-0.14$), and gain reliably from no preparation method. LS even lowers the score of GLM ($-0.12$, $p=0.048$). On Waterbirds with ResNet-18, ERT is ahead of ERM for four of the five editors ($+0.09$ to $+0.23$; not for GLM), so a prepared network should be evaluated with the editor that will work on it.

\paragraph{Edit complexity.}
Without preparation, Astra and Opus already edit CIFAR-100 relabeling and PACS well (Astra with ResNet-18: 0.64 and 0.77; Opus: 0.68 to 0.81), and on these four settings the average gain of every preparation method for Opus is close to zero ($-0.02$ to $+0.03$).

Waterbirds is the hardest benchmark under ERM-trained networks (Astra: 0.40 with ResNet-18 and 0.50 with ViT; Opus: 0.41 and 0.33). ERT raises these scores by $+0.23$ and $+0.24$ for Astra ($p\le0.02$) and by $+0.16$ ($p=0.13$) and $+0.44$ ($p=0.008$) for Opus, ENN by $+0.06$ and $+0.23$, and $+0.09$ and $+0.46$. On the similarly hard Entity-30 confusion fix (Astra: 0.46; Opus: 0.44), ERT gains $+0.33$ and $+0.30$, and ENN $+0.36$ and $+0.16$ (all $p\le0.03$). On the CIFAR-100 confusion fixes, only ERT with ViT gains reliably, for Opus ($+0.12$, $p=0.02$), although it rehearses only relabeling.

For both Astra and Opus, the gain from preparation falls with the score of the standard network (rank correlation over the nine settings for ERT, ENN and LS: $-0.83$, $-0.62$ and $-0.47$ for Astra; $-0.82$, $-0.78$ and $-0.60$ for Opus). Measured on disjoint random halves of the edit requests, so that noise in the standard score cannot create the relation (2{,}000 splits), it remains for ERT and ENN (medians $-0.65$ and $-0.50$ for Astra, $-0.67$ and $-0.53$ for Opus, negative in at least 98\% of splits) and is weaker for LS ($-0.28$ and $-0.35$, negative in 88\% and 92\%). In practice, a few agent edits on the standard network show whether preparation is worth it.

\subsection{Ablations}
\label{sec:ablations}
We have run four main ablations to investigate ERT's dependence on type, location and number of edits, as well as seed used during training.

\paragraph{Edit rule ablation.} To obtain insights on the effect of edit rules used during training, we retrained ERT on Waterbirds with ResNet-18 with seven edit variants: more units, larger steps, a bias shift or sparse edits of the size of Opus's own (see \cref{sec:editpatterns}), ENN's gradient step as additional rule, and stronger rehearsal. No variant differs reliably from the reference for Opus (differences from $-0.06$ to $+0.06$, none significant), and stronger rehearsal, trained for eight of the nine settings (all but the Entity-30 confusion fix), is ahead in three and behind in five. The benefit of ERT thus does not seem to depend on matching the editor's edit closely (details in \cref{sec:supp-rehearse}).

\begin{figure}[t]
  \centering
  \includegraphics[width=\columnwidth]{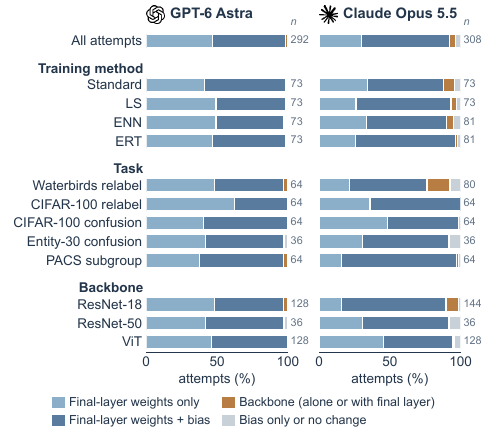}
  \caption{\textbf{Edit locations.} Edits changing only the weights of the final layer, its weights and bias, the backbone (alone or with the final layer), or only the bias or nothing, for the $n$ edit processes of Astra and Opus behind \cref{tab:main}, in total and split by training method, task and backbone.}
  \label{fig:edit-types}
\end{figure}

\paragraph{Backbone edit ablation.} All of the tested rules edit only the final layer, whereas Opus sometimes also edits the backbone (see \cref{fig:edit-types}). To investigate if ERT should facilitate also backbone edits, we retrained a modified ERT on Waterbirds with ResNet-18, where most backbone edits occur. During this training we introduced a third rule: lower the input of the last ReLU on the selected channels, the backbone edit Opus makes most often on ERM trained networks. While the network keeps its accuracy (92.6\% against 92.9\%) under this edit, Opus does not profit: its net score is 0.61 against 0.73 for default ERT on the same eight species (difference $-0.12$, ahead on 2 of 8, $p=0.13$). Although, in this scenario Opus edits the backbone in 3 of 8 edit processes against 1 of 8, this indicates no added benefits.

\begin{figure}[b]
\centering
\includegraphics[width=\columnwidth]{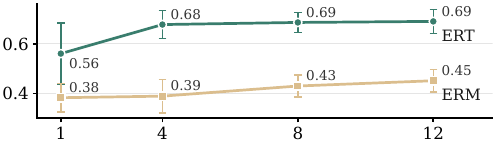}
\vspace{-14pt}
\caption{\textbf{Edit budget ablation.} Opus net score (vertical) on ERM and ERT trained ResNet-18 on Waterbirds for 1 to 12 allowed edit calls (horizontal; bars: $\pm1$ SE). Beyond a budget of 8 the net score does not increase detectably.}
\label{fig:budget}
\vspace{-4pt}
\end{figure}

\paragraph{Training seed ablation.} As most cells of \cref{tab:main} rest on a single trained network per method, it is crucial that the training methods have a low training seed variance. To check this, we use the 26 cells of \cref{tab:main} that have several trained networks per method (PACS, the Entity-30 confusion fix, and ENN and ERT on Waterbirds ResNet-18 with Opus), in which every edit request is run on each network. The mean net score of a network varies between networks by a standard deviation of $0.00$ (95\% CI upper bound $0.03$), whereas edit processes on the same network vary by $0.17$.

\paragraph{Edit budget ablation.} Our weight-space programming protocol allows a maximum of 8 edit calls (see \cref{sec:interface}), and the editor knows this limit from the start. To investigate the impact of the maximum number of allowed edits, we let Opus repeat the Waterbirds relabeling requests on ERM and ERT trained ResNet-18 with budgets of 1, 4, 8 and 12 edits (\cref{fig:budget}). Against 8 calls in the same pilot, neither 4 nor 12 calls change the net score detectably (pooled over both networks $-0.01$, $p=0.86$, and $+0.03$, $p=0.43$; 16 pairs each). Only a single call is lower ($-0.10$, $p=0.08$). Even with 12 calls allowed, Opus uses a median of 5 to 6. ERT is ahead of ERM at every budget, by 0.18 with one call and 0.24 to 0.29 with 4 to 12, and this gap does not depend on the budget (all interactions $p\geq0.36$).

\subsection{LLM Edit Patterns}
\label{sec:editpatterns}

To obtain insights into the edit patterns of LLMs, we analyzed the submitted weight changes of the edit processes behind \cref{tab:main}, 292 of Astra and 308 of Opus. As detailed above, edits on ERM-trained networks observed in early runs motivated the edits that ERT rehearses (see \cref{sec:ert}). The observed edit processes confirm the pre-study observations: in 69 of its 73 edit processes on ERM-trained networks Opus edits the head, with a rank-one change along the contrast between support and reference images, and the paragraphs below show that this shape holds for all tasks, architectures and training methods.

\begin{figure}[b]
\centering
\includegraphics[width=\columnwidth]{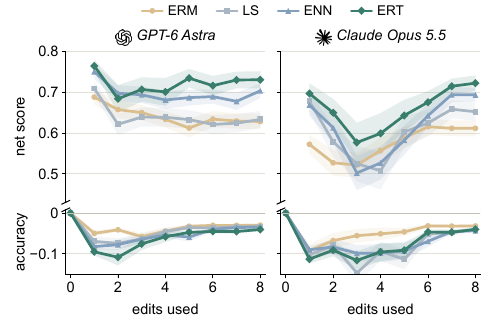}
\vspace{-14pt}
\caption{\textbf{Progress per edit.} Net score (upper curves) and change in test accuracy (lower curves, as a fraction) after each of the 8 edit calls, averaged over the nine settings and all edit processes of \cref{tab:main}. Bands show $\pm 1$ standard error.}
\label{fig:steps}
\vspace{-4pt}
\end{figure}

\paragraph{Edit progression.} To obtain insights on the progress during an edit process, we rebuilt the weights after every edit call from the logged tool calls and scored them like a submission (all 308 edit processes of Opus and 292 of Astra). Both editors reach most of their final net score with the first edit, which also costs the most accuracy on the full test sets (about $-10$ points for Opus, $-5$ to $-10$ for Astra). The later edits repair a substantial share of this side effect (see \cref{fig:steps}). 

Astra trades some success for fewer side effects after the first edit (ERT from 0.77 to 0.73, ERM from 0.69 to 0.63), whereas Opus passes through clearly worse intermediate states (0.50 to 0.58) before it recovers. ERT is ahead of ERM after every edit, by 0.04 to 0.13 for Opus and 0.03 to 0.12 for Astra, so the comparison does not depend on when an edit process is stopped.

For all training methods, accuracy on the full test sets still ends 3 to 4 points below the unedited network after the final edit. Even with frontier LLMs, targeted edits by weight-space programming thus still can cost overall accuracy.

\paragraph{Edit locations.} \cref{fig:edit-types} shows what the edit processes behind \cref{tab:main} change. Both editors almost always edit the final layer: Astra changes its weights in 291 of 292 edit processes and Opus in 296 of 308, in 151 and 199 of them together with its bias. Edits to the backbone are rare, 4 edit processes of Astra and 14 of Opus, and all but one of them change the batch-norm bias before the last ReLU of a ResNet-18. For Opus they concentrate on Waterbirds with ResNet-18 (12 of 48 edit processes, against 2 of 260 elsewhere), and on networks that were not prepared for editing: 5 of 8 edit processes on ERM networks and 2 of 8 with LS, against 4 of 16 on ENN and 1 of 16 on ERT networks.

\paragraph{Edit shapes.} In every task, dataset and architecture, Opus's change to the final layer is rank one (median share of its energy in the first singular value above 99\% in every setting), confined to the rows of the source and the target class, and sparse (median 11 to 64 feature columns, 2--8\% of the features). The columns are chosen by contrast. Of the eight features that differ most between the concept and the reference images, the edit contains 3.2 to 4.5, against 0.1 to 1.4 by chance. The training method does not seem to change this shape.

Among closed-form edits computed from the same support images, the closest to Opus's edits, on the entries it changed, is a rank-one re-route along the contrast, ERT's \emph{direction} rule without removal (mean cosine 0.72 to 0.93 on eight of the nine settings), followed by a gradient step, the edit ENN rehearses (0.51 to 0.87); on the Entity-30 confusion fix, neither is close (0.38 and 0.40). An LLM that edits by hand thus arrives at a sparse version of the edits that both edit-aware methods rehearse, and how gradient-like an edit is does not seem to predict which method wins.

\paragraph{Method differences.} What differs between training methods is the size of the edit. On Waterbirds with ResNet-18, Opus's change to the classifier weights is about five times larger on ERM-trained networks than on prepared ones (median $\ell_2$ norm 7.5, against 1.4 for ERT and 1.6 for ENN), with a similar number of changed units. Preparation thus does not change what kind of edit Opus makes, as it makes a small, ordinary edit sufficient where an ERM-trained network needs a large one or an edit inside the backbone.

\section{Conclusion and Future Work}
\label{sec:conclusion}

We presented the first systematic study of weight-space programming for vision models, with a benchmark of LLM agents that edit trained classifiers and with Edit-Ready Training (ERT), which rehearses the head edits these agents write. Over more than 2{,}600 scored edit processes, ERT has the highest mean net score for both GPT-6 Astra and Claude Opus 5.5. Further, standard networks already support targeted edits in easy settings, and preparation pays off where they resist.

Our study also leaves several directions open. It covers image classification with three backbones and four datasets. Other vision tasks, larger backbones and generative models remain to be tested. With 8 to 16 paired edit processes per cell, only large differences between methods are detectable. Hyperparameters were tuned and baselines selected with scripted editors on ResNet-18 rather than with agents. Finally, with every training method, edits still lower test accuracy. Preparation that also preserves accuracy, rehearsal of edits inside the backbone or tailored to a specific editor, and sequences of edits to one network are natural next steps. We release all tasks, trajectories and scores so that new methods and editors can be compared directly.

\section*{Acknowledgments}
The authors acknowledge support by the state of Baden-W\"urttemberg through bwHPC.

\section*{LLM Usage Statement}
\label{sec:llm-usage}

The authors take full responsibility for all submitted content, regardless of any LLM or AI tool use. Beyond the LLM editors that are the object of study, we used Claude Code (Opus and Fable) as well as GPT-6 Astra as described below.

\noindent\textbf{Ideas and conclusions.} The research questions, the method and the conclusions are the authors' own. Claude was used to discuss and sharpen their formulation.

\noindent\textbf{Related work and writing.} Claude proposed related literature, from which the authors selected the cited works and checked each reference against its source. Claude and GPT-6 edited text for grammar, clarity and structure and provided LaTeX code for tables, while the authors approved all edits.

\noindent\textbf{Code, experiments and figures.} Claude Code implemented the method, the tool interface and the analysis with limited step-by-step direction, launched and monitored the training and editing runs, and generated the result tables. The authors specified the method, tasks, protocol and metrics, performed code checks, verified the reported numbers, and interpreted the results.

{\small
\bibliographystyle{ieeenat_fullname}
\bibliography{bench}
}

\clearpage
\appendix
\raggedbottom

\noindent This appendix provides the details that the main text refers to. It is organized as follows:
\begin{itemize}\setlength\itemsep{0pt}
\item \cref{sec:supp-protocol}: the WeightLab tools, the system prompt and task objectives given to the editors, and how the five editors were run.
\item \cref{sec:supp-training}: training settings, hyperparameter search spaces and selected values, and the baseline screen.
\item \cref{sec:supp-eval}: edit requests, training seeds and statistical procedures.
\item \cref{sec:supp-main}: per-setting results and paired tests for Astra and Opus.
\item \cref{sec:supp-open}: results of the three open-weight editors.
\item \cref{sec:supp-ablations}: ablations of the rehearsed edit, of held-out concepts and of the edit budget.
\item \cref{sec:supp-edits}: further analysis of the submitted edits.
\item \cref{sec:supp-scale}: scale of the study, compute and release.
\end{itemize}
As in the main text, ERM denotes standard training, LS label smoothing, ENN Editable Neural Networks and ERT our Edit-Ready Training.

\section{WeightLab Protocol}
\label{sec:supp-protocol}

This section complements \cref{sec:interface}. Every edit process is served by a fresh WeightLab server process that holds one network and one edit request.

\subsection{Tools and budget}
\label{sec:supp-tools}

\begin{table*}[t]
\centering
\footnotesize
\setlength{\tabcolsep}{6pt}
\begin{tabular}{@{}lll@{}}
\toprule
Tool & Charged as & Function \\
\midrule
\tool{open\_task} & free & Returns the task objective, a summary of the network and the budget \\
\tool{budget\_status} & free & Returns the remaining budget \\
\midrule
\tool{list\_parameters} & read & Names and shapes of the editable tensors \\
\tool{list\_layers} & read & Layers whose activations can be probed, with their number of channels \\
\tool{read\_weights} & read & Values of a tensor slice (at most 512 numbers), or summary statistics per slice \\
\midrule
\tool{probe\_predictions} & probe & Accuracy and mean logit margin on the support and reference images, optionally per image \\
\tool{probe\_activations} & probe & Mean activation of the channels of one layer per image group, ranked by their difference \\
\tool{probe\_logit\_contributions} & probe & Contribution of every feature unit to the margin between two classes, per image group \\
\midrule
\tool{set\_weights} & edit & Overwrites a tensor slice \\
\tool{add\_to\_weights} & edit & Adds values to a tensor slice \\
\tool{scale\_weights} & edit & Multiplies a tensor slice (row, output channel or whole tensor) by a factor \\
\tool{scale\_input\_channels} & edit & Multiplies input-channel slices of a convolutional or linear weight by a factor \\
\tool{edit\_entries} & edit & Changes a list of individual entries of one tensor in a single call \\
\midrule
\tool{undo\_last\_edit} & free & Reverts the most recent edit (the edit stays charged) \\
\tool{reset\_to\_original} & free & Restores the original weights (edits stay charged) \\
\tool{submit} & free & Ends the edit process; the current weights are scored \\
\bottomrule
\end{tabular}
\caption{\textbf{WeightLab tools.} Every edit process is limited to 20 probes, 30 reads and 8 edits. All probes run the current, possibly edited, network on the support and reference images of the edit request only.}
\label{tab:tools}
\end{table*}

\Cref{tab:tools} lists the 16 tools that WeightLab exposes. Each tool is either free or charged to one of the three budgets of 20 probes, 30 reads and 8 edits. A call is validated before it is charged, so a malformed call or an edit of more than 65{,}536 parameters is rejected without cost. Images are never returned. The editor only sees predictions, margins and activations on the support and reference images, which the server calls subgroup~1 and subgroup~0. The server logs every call with its arguments and result, and stores the submitted weights.

\subsection{System prompt}
\label{sec:supp-prompt}

All five editors receive the same system prompt for every edit process, independent of task, dataset, backbone and training method:
\begin{promptbox}{System prompt}{all editors, all edit processes}
You are a careful model-editing engineer. You are given the ID of one editing task. Your job is to improve a trained neural network by directly editing its existing weights, using only the weightlab tools.

\textbf{Rules:}
\begin{itemize}\setlength\itemsep{1.5pt}\setlength\parskip{0pt}\setlength\topsep{0pt}
\item Use only the weightlab tools. Start with \ptool{open\_task(task\_id)} and read the objective and budget carefully.
\item You can only run the model forward on the provided support examples (probes), read weights, and apply direct weight edits. There are no gradients, no training, no added layers or adapters.
\item Probes, reads and edits are budgeted; the server enforces the budget. \ptool{undo\_last\_edit} and \ptool{reset\_to\_original} are free but do not refund budget. Plan before spending.
\item Work from evidence: work out which internal units or weights cause the poor subgroup performance, test it with probes, then make targeted edits and verify their effect.
\item Your final weights are scored on held-out data from the same subgroups. Avoid edits that only fit the support examples.
\item Do not try to find out anything about where the model came from or how it was trained; that information is not available and not needed.
\item Always finish by calling \ptool{submit()} with a concise summary (what you changed, why, and the support-set result). If your edits did not help, \ptool{reset\_to\_original()} before submitting.
\end{itemize}

Your final reply should be a short report: the edits applied, the support-set worst-subgroup and overall accuracy before and after, and your confidence that the edit generalizes.
\end{promptbox}
\noindent The only user message of an edit process is the following, where the ID is a random string that carries no information about the network:
\begin{promptbox}{User message}{all editors}
Task ID: \ptool{\textless id\textgreater}
\end{promptbox}
\noindent In the boxes of this section, only the typesetting is ours (bold face, and the color of tool names); the wording is verbatim.

\subsection{Task objectives}
\label{sec:supp-objectives}

The edit request itself is returned by \tool{open\_task} as the task objective. There is one objective text per task, shown in the three boxes below, in which only the number of classes $C$ varies ($C=2$ on Waterbirds, 20 on CIFAR-100, 30 on Entity-30 and 7 on PACS).

\begin{promptbox}{Task objective: relabel a concept}{Waterbirds, CIFAR-100}
You are given a trained $C$-class image classifier and a small labeled support set with two subgroups. Subgroup 1 contains a few examples of one specific visual concept that the model currently assigns to one class; the labels given for subgroup 1 are the class it SHOULD be assigned to instead. Subgroup 0 is a reference set of unrelated examples, labeled with the model's CURRENT predictions, which must not change. Directly edit the existing weights so that the concept is assigned to its new class, while changing as few other predictions as possible. You are scored on held-out examples of the same concept (fraction assigned to the new class) and on held-out unrelated examples (fraction of predictions left unchanged); both matter. Prefer targeted, principled edits that generalize over edits that merely memorize the support examples. Use forward probes to understand the model before editing. Call \ptool{submit()} when done; if you never submit, the original weights are scored.
\end{promptbox}

\begin{promptbox}{Task objective: fix a confusion}{CIFAR-100, Entity-30}
You are given a trained $C$-class image classifier and a small labeled support set with two subgroups. Subgroup 1 contains examples of one class that the model often confuses with one other class; the labels given for subgroup 1 are their correct class. Subgroup 0 is a reference set of other examples, labeled with the model's CURRENT predictions, which must not change. Directly edit the existing weights so that the model stops making this confusion, while changing as few other predictions as possible. You are scored on held-out examples of the same class that the model currently confuses in this way (fraction now classified correctly) and on held-out other examples (fraction of currently correct predictions that stay correct); both matter. Prefer targeted, principled edits that generalize over edits that merely memorize the support examples. Use forward probes to understand the model before editing. Call \ptool{submit()} when done; if you never submit, the original weights are scored.
\end{promptbox}

\begin{promptbox}{Task objective: fix a subgroup}{PACS}
You are given a trained $C$-class image classifier and a small labeled support set with two subgroups. Subgroup 1 contains examples of one class drawn from a kind of input on which the model performs poorly; the labels given for subgroup 1 are their correct class. Subgroup 0 is a reference set of other examples, labeled with the model's CURRENT predictions, which must not change. Directly edit the existing weights so that the model classifies inputs like subgroup 1 correctly, while changing as few other predictions as possible. You are scored on held-out examples of the same kind and class that the model currently misclassifies (fraction now classified correctly) and on held-out other examples (fraction of currently correct predictions that stay correct); both matter. Prefer targeted, principled edits that generalize over edits that merely memorize the support examples. Use forward probes to understand the model before editing. Call \ptool{submit()} when done; if you never submit, the original weights are scored.
\end{promptbox}

\subsection{How the editors were run}
\label{sec:supp-harness}

Tools, budget, system prompt and user message are identical for all editors. The harnesses that run them differ in three points.

\paragraph{Claude Opus 5.5} runs at default effort as a sub-agent of Claude Code that has access to the WeightLab tools only. It is limited to 150 model turns, where one turn is one model response that may contain several tool calls.

\paragraph{GPT-6 Astra} runs at high reasoning effort in the Codex command-line agent, whose base instructions are replaced by the system prompt above and whose only tools are those of WeightLab. It has no turn limit, and an edit process ends when the model stops. Every transcript was audited for tool use outside WeightLab.

\paragraph{Open-weight editors.} GLM-5.3-Flash and DeepSeek-V4.1-Flash (low reasoning effort) and Gemma-4-26B (thinking mode) run in a chat-completions loop with the same limit of 150 turns and at most 32{,}000 output tokens per turn. Tool outputs longer than 30{,}000 characters are truncated with a notice, so that they fit the context windows of these models.

\paragraph{Continuation reminders.} The open-weight models sometimes end a turn without a tool call before they have submitted. In this case, they receive the neutral reminder
\begin{promptbox}{Continuation reminder}{open-weight editors only}
Continue. Use the tools, and call \ptool{submit()} when you are done.
\end{promptbox}
\noindent at most three times per edit process. After the third reminder, the edit process ends, and the original weights are scored unless the editor has submitted. Astra and Opus receive no reminders. The reminders are logged for every edit process.

\paragraph{Order and exclusions.} The edit processes of one editor ran strictly one at a time, in shuffled order within each grid. The ERM and LS arms of most settings were added one to three days after the ENN and ERT arms of the same edit requests, with the same tool server and prompt; they were not interleaved. An edit process was excluded and rebuilt only after an infrastructure failure before submission, such as a lost connection. A submitted edit process was never excluded.

\section{Training Details}
\label{sec:supp-training}

This section complements \cref{sec:networktraining}.

\paragraph{Common settings.} All networks start from ImageNet weights and are trained with SGD (learning rate $10^{-3}$, momentum 0.9, weight decay $10^{-4}$, gradient-norm clipping at 5). Each training batch holds 64 images. ENN and ERT add four episodes of four support and four query images, and the reference images of an episode are the batch images of other concepts. On the multi-class datasets, ERT and ENN draw a random target class per episode.

\paragraph{Held-out concepts.} The concepts that the editors later relabel are excluded from the training episodes of ENN and ERT, on Waterbirds (8 species) and on CIFAR-100 (16 fine classes). On PACS, the (class, sketch) cells that are fixed later are excluded. The images of these concepts remain in the training set with their original labels. \Cref{sec:supp-heldout} measures how much this exclusion matters.

\subsection{Hyperparameter tuning}
\label{sec:supp-tuning}

\begin{table}[t]
\centering
\footnotesize
\setlength{\tabcolsep}{4pt}
\begin{tabular}{@{}lll@{}}
\toprule
Method & Hyperparameter & Search space \\
\midrule
ERT & edit weight $\lambda_e$ & $\{0.3,\,1,\,3\}$ \\
 & reference weight $\lambda_l$ & $\{1,\,5\}$ \\
 & warm-up (epochs) & $\{0,\,2\}$ \\
\midrule
ENN & step size $\eta$ & $\{0.1,\,0.5,\,2\}$ \\
 & gradient steps & $\{1,\,5\}$ \\
 & reference weight $\lambda_l$ & $\{0.1,\,1\}$, then $\{5,\,20\}$ \\
\midrule
LS & smoothing strength & $\{0.05,\,0.1,\,0.2\}$ \\
\midrule
Dropout & rate & $\{0.1,\,0.3,\,0.5\}$ \\
SAM & radius $\rho$ & $\{0.02,\,0.05,\,0.2\}$ \\
Concept labels & auxiliary loss weight & $\{0.3,\,1,\,3\}$ \\
Sparse final layer & elastic-net strength & 40-point path \\
\bottomrule
\end{tabular}
\caption{\textbf{Search spaces.} All combinations were trained with ResNet-18 on Waterbirds and CIFAR-100, on two training seeds that are used for nothing else. The lower block lists the methods that were only part of the baseline screen (\cref{sec:supp-screen}).}
\label{tab:search}
\end{table}

\begin{table}[t]
\centering
\footnotesize
\setlength{\tabcolsep}{3pt}
\begin{tabular}{@{}lll@{}}
\toprule
Setting & ERT & ENN \\
\midrule
Waterbirds, 100 ep. & $\lambda_e=3$, $\lambda_l=5$, & $\eta=0.5$, 1 step, $\lambda_l=1$ \\
 & no warm-up & (ViT-B/16: $\lambda_l=5$) \\
CIFAR-100, 30 ep. & $\lambda_e=\lambda_l=1$, 2-ep. & $\eta=0.5$, 5 steps, $\lambda_l=1$, \\
 & warm-up, LS 0.2 & LS 0.2 \\
PACS, 50 ep. & as CIFAR-100 & as CIFAR-100 \\
Entity-30, 20 ep. & as CIFAR-100, & as CIFAR-100, \\
 & without LS & without LS \\
\bottomrule
\end{tabular}
\caption{\textbf{Selected settings.} LS uses strength 0.2 everywhere. ViT-B/16 networks reuse the ResNet-18 settings of their dataset, and PACS and Entity-30 reuse the settings selected on CIFAR-100.}
\label{tab:settings}
\end{table}

All methods were tuned with the same protocol, fixed in advance, with ResNet-18 on Waterbirds and on CIFAR-100. \Cref{tab:search} lists the search spaces. Every combination was trained on two training seeds (5 and 6) that appear in no reported result, and was scored on validation data with three scripted editors: the two ERT edit rules and a single gradient step on the head. The selection score is the mean success minus side effect of these editors on validation concepts. Among the settings whose validation accuracy is within two points of ERM, the one with the highest score was selected. No LLM was involved in tuning.

\Cref{tab:settings} lists the selected values. On CIFAR-100, both ERT and ENN are combined with LS, since this improved both in the same protocol. For ENN on Waterbirds, the initially selected $\lambda_l=1$ lay at the upper end of its range. We therefore extended the range to $\{5,20\}$, which selected $\lambda_l=5$. The ResNet-18 network had been trained before this extension and uses $\lambda_l=1$, whereas the ViT-B/16 network uses $\lambda_l=5$; with scripted editors, the two values differ by 0.01 in mean net score. For ERT, an extended grid on Waterbirds ($\lambda_e$ up to 10, $\lambda_l$ from 0.2 to 20) left the selection unchanged.

\subsection{Baseline screen}
\label{sec:supp-screen}

\begin{table}[t]
\centering
\footnotesize
\setlength{\tabcolsep}{3.5pt}
\begin{tabular}{@{}lccccc@{}}
\toprule
 & \multicolumn{2}{c}{Waterbirds} & \multicolumn{2}{c}{CIFAR-100} & \\
\cmidrule(lr){2-3}\cmidrule(lr){4-5}
Method & 20 ep. & 100 ep. & 20 ep. & 100 ep. & Gain \\
\midrule
ERM & 0.39 & 0.36 & 0.38 & 0.36 & -- \\
Dropout & 0.39 & 0.36 & 0.39 & 0.37 & $+0.00$ \\
SAM & 0.40 & 0.36 & 0.41 & 0.37 & $+0.01$ \\
Concept labels & 0.56 & 0.55 & 0.38 & 0.36 & $+0.09$ \\
Sparse final layer & 0.57 & 0.53 & 0.46 & 0.47 & $+0.13$ \\
LS & 0.51 & 0.51 & 0.57 & 0.65 & $+0.19$ \\
ERT & 0.66 & \textbf{0.76} & 0.51 & 0.45 & $+0.22$ \\
ENN & \textbf{0.69} & 0.70 & 0.50 & 0.56 & $+0.24$ \\
\midrule
ERT + LS & -- & -- & 0.59 & 0.61 & -- \\
ENN + LS & -- & -- & \textbf{0.63} & \textbf{0.70} & -- \\
\bottomrule
\end{tabular}
\caption{\textbf{Baseline screen with scripted editors.} Mean net score of six scripted editors on ResNet-18 networks, averaged over five training seeds, at an early and at the last checkpoint. \emph{Gain}: difference to ERM, averaged over datasets and checkpoints. No LLM is involved. Bold: best method of a column.}
\label{tab:screen}
\end{table}

LLM edit processes are expensive, so the baselines of the main text were chosen by a screen without LLMs (\cref{tab:screen}). Ten training methods were tuned as described above and trained with ResNet-18 on Waterbirds and CIFAR-100 over five training seeds. Each network was then edited by six scripted editors on the test concepts: the two ERT edit rules, three sparse refits of the head that are restricted to 1, 4 and 16 feature units, and a rescaling of 32 channels in the last block of the backbone.

Sharpness-aware minimization and dropout are indistinguishable from ERM in this screen. Auxiliary concept labels and a sparse final layer gain $+0.09$ and $+0.13$, below LS ($+0.19$), ERT ($+0.22$) and ENN ($+0.24$). We therefore gave the three leading methods and ERM to the LLM editors, with LS added to ERT and ENN where it was selected. The Waterbirds networks of this screen were trained before the species were held out from the episodes (\cref{sec:supp-heldout}), so the screen was used only for selecting the baselines.

\section{Evaluation Details}
\label{sec:supp-eval}

This section complements \cref{sec:weightspaceprogramming}.

\paragraph{Edit requests and training seeds.} \emph{Waterbirds:} 8 bird species, training seed 0; for Opus with ResNet-18, ENN and ERT are also edited on seed 1. \emph{CIFAR-100:} 8 of the 16 held-out fine classes for relabeling, and the 8 superclass pairs with the highest validation confusion for the confusion fix, seed 0. \emph{PACS:} the 4 (class, sketch) cells with the lowest validation accuracy, seeds 0 and 1. \emph{Entity-30:} the 3 class pairs with the highest validation confusion, seeds 0 to 2.

\paragraph{Support and reference images.} The support set holds five images of the concept for relabeling, and eight images of the target group, labeled with their correct class, for the two fix tasks. The reference set holds 64 validation images of other concepts, labeled with the predictions of the respective network. All training methods of a setting share the same support and reference images.

\paragraph{Intervals and tests of the main table.} The intervals in \cref{tab:main} are nonparametric percentile bootstrap intervals over the edit processes of a cell (10{,}000 resamples); the mean column resamples within each setting. A cell has 8 edit processes, except on Entity-30 (9) and for ENN and ERT on Waterbirds with ResNet-18 and Opus (16, over two training seeds). The paired sign-flip test is exact for up to 18 pairs. With 8 pairs, its smallest attainable $p$-value is 0.008, and with 9 pairs 0.004.

\paragraph{Tests of the mean gain.} For the mean gain over ERM, we report two tests. The first uses all 73 pairs of edit processes of an editor, flips signs within pairs and weights the nine settings equally (20{,}000 random flips). It tests whether a method changes the outcome on these nine settings. The second takes the nine settings as units (exact sign-flip test over the setting means) and asks whether the gain would carry over to new settings.

\paragraph{Split-half check (\cref{sec:main}).} For each setting, the paired edit requests are split at random into two halves. The net score on the ERM network is taken from one half and the gain (method minus ERM) from the other. The rank correlation over the nine settings is computed for 2{,}000 random splits, and we report its median and the share of negative values. Because the two halves share no edit processes, noise in the ERM score cannot create the correlation.

\section{Additional Results for Astra and Opus}
\label{sec:supp-main}

\begin{table*}[t]
\centering
\small
\setlength{\tabcolsep}{5pt}
\begin{tabular}{@{}l ccc ccc ccc@{}}
\toprule
& \multicolumn{3}{c}{LS $-$ ERM} & \multicolumn{3}{c}{ENN $-$ ERM} & \multicolumn{3}{c}{ERT $-$ ERM} \\
\cmidrule(lr){2-4}\cmidrule(lr){5-7}\cmidrule(lr){8-10}
Setting & diff. & wins & $p$ & diff. & wins & $p$ & diff. & wins & $p$ \\
\midrule
\multicolumn{10}{@{}l}{\textit{GPT-6 Astra}} \\
\inputtab tables_bench/tests_astra.tex
\midrule
\multicolumn{10}{@{}l}{\textit{Claude Opus 5.5}} \\
\inputtab tables_bench/tests_opus.tex
\bottomrule
\end{tabular}
\caption{\textbf{Paired tests against ERM.} Mean paired difference in net score, number of pairs in which the method is ahead of ERM, and exact two-sided sign-flip $p$-value, for every setting of \cref{tab:main}. On Waterbirds with ResNet-18, the ERM and LS networks have one training seed, so Opus's ENN and ERT edit processes are paired on that seed only (8 pairs).}
\label{tab:tests}
\end{table*}

\begin{table*}[t]
\centering
\footnotesize
\setlength{\tabcolsep}{3pt}
\resizebox{\textwidth}{!}{%
\begin{tabular}{@{}l cccc cccc cccc cccc@{}}
\toprule
& \multicolumn{4}{c}{ERM} & \multicolumn{4}{c}{LS} & \multicolumn{4}{c}{ENN} & \multicolumn{4}{c}{ERT} \\
\cmidrule(lr){2-5}\cmidrule(lr){6-9}\cmidrule(lr){10-13}\cmidrule(lr){14-17}
Setting & succ. & side & $\Delta$acc & acc & succ. & side & $\Delta$acc & acc & succ. & side & $\Delta$acc & acc & succ. & side & $\Delta$acc & acc \\
\midrule
\inputtab tables_bench/detail_opus.tex
\bottomrule
\end{tabular}}
\caption{\textbf{Success, side effect and accuracy for Opus.} \emph{succ.}\ and \emph{side}: success and side effect as defined in \cref{sec:weightspaceprogramming}, whose difference is the net score. $\Delta$acc: change in test accuracy caused by the edit, in percentage points. \emph{acc}: test accuracy (\%) of the network before editing (not logged for the CIFAR-100 confusion fix with ViT-B/16). On relabeling tasks, accuracy is measured against the original labels, so the relabeled concept itself counts as an error after a successful edit; it is a small part of the test set.}
\label{tab:detail}
\end{table*}

\begin{table}[t]
\centering
\footnotesize
\setlength{\tabcolsep}{4pt}
\begin{tabular}{@{}lccc@{}}
\toprule
Setting & ERT $-$ ENN & wins & $p$ \\
\midrule
\inputtab tables_bench/ert_vs_enn.tex
\bottomrule
\end{tabular}
\caption{\textbf{ERT against ENN for Opus}, paired on the same edit requests (Waterbirds with ResNet-18: two training seeds).}
\label{tab:ertenn}
\end{table}

\paragraph{Paired tests.} \Cref{tab:tests} lists the paired comparisons against ERM behind \cref{tab:main} for both main editors, and \cref{tab:ertenn} the direct comparison of ERT and ENN for Opus. Over all 73 pairs of edit processes (\cref{sec:supp-eval}), the mean gain over ERM is $+0.12$ for ERT ($p<0.001$), $+0.08$ for ENN ($p=0.007$) and $+0.05$ for LS ($p=0.08$) with Opus, and $+0.12$ ($p<0.001$), $+0.08$ ($p=0.002$) and $+0.00$ ($p=0.88$) with Astra. With the nine settings as units, the $p$-values for Opus are 0.047 (ERT), 0.17 (ENN) and 0.23 (LS).

\paragraph{Success, side effect and accuracy.} \Cref{tab:detail} splits every net score of Opus into success and side effect, and adds the change in test accuracy caused by the edit as well as the test accuracy of the network before editing. The networks of the four training methods are similarly accurate before editing, within 1 to 2 points on CIFAR-100 and Entity-30. On Waterbirds, ENN and ERT are 4 to 7 points more accurate than ERM and LS, and on PACS with ViT-B/16, ERT is 4 to 5 points less accurate than the other methods.

\paragraph{Agreement of the two main editors.} Over the 36 cells of \cref{tab:main}, the rank correlation between the net scores of Astra and Opus is $0.78$, and the rank correlation between their gains from preparation (method minus ERM, per setting; 27 gains) is $0.77$.

\section{Open-Weight Editors}
\label{sec:supp-open}

\begin{table*}[t]
\centering
\small
\setlength{\tabcolsep}{2pt}
\renewcommand{\arraystretch}{1.05}
\renewcommand{\cistack}[2]{#1\,{\fontsize{5.5}{6.5}\selectfont\color{black!55}[#2]}}
\caption{\textbf{Main results for the open-weight editors.}
Net score ($\uparrow$) with 95\% bootstrap confidence interval of GLM-5.3-Flash, Gemma-4-26B and DeepSeek-V4.1-Flash edits on networks trained in four ways, in the layout of \cref{tab:main}. WB=Waterbirds, C100=CIFAR-100; R18=ResNet-18; LS=label smoothing. On CIFAR-100 and PACS, ENN and ERT include label smoothing. Bold: best method of a column, underlined: second best. None of the three edited the Entity-30 confusion fix, so its column is omitted and the mean covers the eight settings shown. Every cell has 8 edit processes. $\Delta$ rows: paired difference between ERT and ERM, and between ERT and the best baseline, with $p<0.05$ (paired sign-flip test) in bold.}
\label{tab:others}
\inputtab tables_bench/table1_open.tex
\end{table*}

\begin{figure*}[t]
\centering
\includegraphics[width=\textwidth]{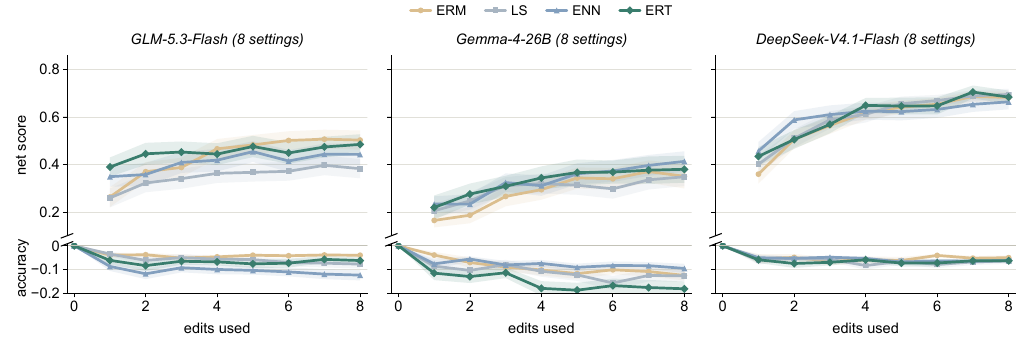}
\vspace{-14pt}
\caption{\textbf{Progress per edit for the open-weight editors.} Net score (upper curves) and change in test accuracy (lower curves, as a fraction) after each of the 8 edit calls, as in \cref{fig:steps}. Each editor is averaged over the settings it completed with all four methods (number in the panel title). Both quantities use the same scale, and the axis is cut in between. An edit process that stops early keeps its last state. Bands show $\pm 1$ standard error.}
\label{fig:steps-open}
\end{figure*}

\Cref{tab:others} repeats \cref{tab:main} for the three open-weight editors. Each of them covers eight of the nine settings with all four training methods.

\paragraph{Agreement with Opus.} Over the cells that both edited, the rank correlation of the net score with Opus is $0.02$ for GLM, $-0.14$ for Gemma and $0.65$ for DeepSeek (32 cells each). The rank correlation of the gains from preparation is $0.14$ for GLM, $-0.19$ for Gemma and $0.31$ for DeepSeek (24 gains each).

\paragraph{Gains from preparation.} The mean paired gain over ERM, with the settings weighted equally and tested as for Opus (\cref{sec:supp-eval}), is $-0.12$ for LS ($p=0.048$), $-0.06$ for ENN ($p=0.33$) and $-0.01$ for ERT ($p=0.90$) with GLM, and $-0.09$ ($p=0.12$), $-0.03$ ($p=0.57$) and $+0.06$ ($p=0.30$) with Gemma, and $+0.03$ ($p=0.34$), $-0.01$ ($p=0.74$) and $+0.03$ ($p=0.44$) with DeepSeek (64 pairs over eight settings each).

\paragraph{Progress per edit.} \Cref{fig:steps-open} repeats the progress-per-edit analysis of the main text. Unlike Astra and Opus, the open-weight editors do not reach their final net score with the first edit. It rises over the whole budget, from 0.26--0.39 to 0.38--0.50 for GLM, from 0.17--0.23 to 0.35--0.41 for Gemma and from 0.36--0.46 to 0.66--0.70 for DeepSeek (range over the four training methods). They also do not win back the accuracy lost with the first edits: GLM ends 4 to 12 points and DeepSeek 5 to 6 points below the unedited network. Gemma, which uses all 8 edit calls in 57\% of its edit processes, keeps losing accuracy as it edits and ends 10 to 18 points below it.

\section{Ablations}
\label{sec:supp-ablations}

This section complements \cref{sec:ablations}. Unless stated otherwise, the ablations use Waterbirds with ResNet-18 and Opus as editor, under the protocol of the main grid. Their networks were trained before the edited species were held out from the training episodes (\cref{sec:supp-heldout}). Reference and variant share this property in every comparison below.

\subsection{Which edit should ERT rehearse?}
\label{sec:supp-rehearse}

\begin{table}[t]
\centering
\footnotesize
\setlength{\tabcolsep}{4pt}
\inputtab tables_bench/ablation_edits.tex
\caption{\textbf{Which edit should ERT rehearse?} Net score of Opus on Waterbirds with ResNet-18 when the edit simulated during training is varied and everything else is kept. $\Delta$ is the paired difference to the reference of the same block on the same species, with the number of pairs won and the sign-flip $p$-value. The two blocks were run separately, each with its own reference edit processes. No variant differs reliably from the reference.}
\label{tab:ablation-edits}
\end{table}

\begin{table*}[t]
\centering
\small
\setlength{\tabcolsep}{5pt}
\inputtab tables_bench/ablation_edits_settings.tex
\caption{\textbf{Rehearsed-edit variants in further settings.} Net score of Opus with reference ERT and with the variant, paired on the same edit requests, with the mean paired difference $\Delta$ and the sign-flip $p$-value.}
\label{tab:ablation-settings}
\end{table*}

ERT rehearses two closed-form edit rules, and one may expect that rehearsing edits closer to what the LLM later writes would help more. We varied the simulated edit on Waterbirds with ResNet-18, the setting with the largest gain from preparation at that time, retrained the network and gave it to Opus (\cref{tab:ablation-edits}).

\paragraph{Variants.} Opus's own edits change more units than the top-8 rule, shift the class bias and often overshoot the margin (\cref{sec:editpatterns}). The variants therefore enlarge or randomize the number of units, scale the step size by a random factor between 1 and 3, shift the bias so that the mean reference logits stay unchanged, or add sparse edits with the size statistics of Opus's edits to the two rules. A further variant adds ENN's gradient step as a third rule, and \emph{stronger rehearsal} draws 16 instead of 4 episodes per step, marks 64 units and uses two support images.

\paragraph{Results.} No variant differs reliably from the reference: the differences range from $-0.06$ to $+0.06$, and none is significant. Two variants that moved furthest toward Opus's edits could not be tested at all, because they lowered the accuracy of the unedited network (for an edit fitted directly to Opus's edits, 88.9\% against 92.9\%).

Stronger rehearsal, the only variant with the same gain on both training seeds, was then trained for eight of the nine settings, all but the Entity-30 confusion fix (\cref{tab:ablation-settings}). It is ahead of the reference in three of them and behind in five (mean difference $-0.03$), and the only significant difference is a gain on PACS with ResNet-18 ($+0.07$, 7 of 8 pairs, $p=0.02$). On ViT-B/16, where about half of the entries Opus changes are negative, we also replaced the top-8 rule by a signed sparse edit. It changed the net score by $-0.01$ on Waterbirds and by $+0.03$ on CIFAR-100 relabeling (both $p>0.3$).

The benefit of ERT thus does not depend on matching the editor's edit closely. Together with the finding that Opus writes the same kind of edit on every network (\cref{sec:editpatterns}), this suggests that rehearsal shapes the features so that small head edits work in general, and not one particular edit.

\subsection{Held-out concepts}
\label{sec:supp-heldout}

\begin{table}[t]
\centering
\footnotesize
\setlength{\tabcolsep}{4pt}
\begin{tabular}{@{}ll cc cc@{}}
\toprule
& & \multicolumn{2}{c}{ENN} & \multicolumn{2}{c}{ERT} \\
\cmidrule(lr){3-4}\cmidrule(lr){5-6}
Editor & Backbone & rehearsed & held out & rehearsed & held out \\
\midrule
Astra & ResNet-18 & 0.52 & 0.47 & 0.57 & 0.63 \\
 & ViT-B/16 & 0.71 & 0.74 & 0.78 & 0.74 \\
Opus & ResNet-18 & 0.58 & 0.52 & 0.72 & 0.63 \\
 & ViT-B/16 & 0.83 & 0.79 & 0.82 & 0.76 \\
\midrule
GLM & ResNet-18 & 0.59 & 0.30 & 0.52 & 0.38 \\
 & ViT-B/16 & 0.62 & 0.53 & 0.26 & 0.24 \\
Gemma & ResNet-18 & 0.16 & 0.23 & 0.15 & 0.30 \\
 & ViT-B/16 & 0.23 & 0.12 & 0.29 & 0.20 \\
DeepSeek & ResNet-18 & 0.59 & 0.52 & 0.55 & 0.64 \\
 & ViT-B/16 & 0.53 & 0.75 & 0.78 & 0.62 \\
\bottomrule
\end{tabular}
\caption{\textbf{Held-out concepts on Waterbirds.} Net score when the edited species were part of the training episodes (\emph{rehearsed}) and when they were excluded from them (\emph{held out}), for the same edit requests. Cells have 8 edit processes (16 for Opus with ResNet-18). The main text reports the held-out networks.}
\label{tab:heldout}
\end{table}

Edit episodes may only use concepts that are not edited later (\cref{sec:ert}). To measure how much this matters, we compare two versions of the Waterbirds ENN and ERT networks that differ only in this point. In the \emph{rehearsed} version, the 8 species that are relabeled later, each with 22 to 26 training images, took part in the training episodes. The test images that score an edit were unseen, but the concept and its target class were not. In the \emph{held-out} version, which the main text reports, these species are excluded from the episodes. Nothing was retuned, and the accuracy of the unedited networks differs by about one point at most.

\Cref{tab:heldout} shows both versions for all five editors. For Opus, holding out the species lowers the net score of ERT by 0.08 over both backbones (24 pairs, $p=0.02$) and that of ENN by 0.05 ($p=0.24$). For Astra, neither changes (ERT $+0.01$, $p=0.86$; ENN $-0.01$, $p=0.65$; 16 pairs each). The gain of ERT over ERM remains with held-out species, with $+0.29$ for Opus (24 pairs, $p<0.001$; edit processes on training seed 1 are paired with the ERM edit process of the same species) and $+0.23$ for Astra (16 pairs, $p<0.001$) over both backbones. A part of the benefit of ERT for Opus on Waterbirds thus comes from having rehearsed the concept itself, and the larger part carries over to concepts that were never rehearsed. For the open-weight editors, the differences between the two versions have both signs and reflect the high variance of these editors.

\subsection{Edit budget}
\label{sec:supp-budget}

\begin{table}[t]
\centering
\footnotesize
\setlength{\tabcolsep}{5pt}
\begin{tabular}{@{}lc cc cc@{}}
\toprule
& & \multicolumn{2}{c}{Net score} & \multicolumn{2}{c}{Edit calls used} \\
\cmidrule(lr){3-4}\cmidrule(lr){5-6}
& Budget & ERM & ERT & ERM & ERT \\
\midrule
\emph{First run} & 1 & 0.38 & 0.56 & 1 & 1 \\
 & 2 & 0.47 & 0.66 & 2 & 2 \\
 & 8 & 0.46 & 0.69 & 6 & 6 \\
\midrule
\emph{Second run} & 4 & 0.39 & 0.68 & 4 & 4 \\
 & 8 & 0.40 & 0.68 & 6 & 6 \\
 & 12 & 0.45 & 0.69 & 5 & 6 \\
\bottomrule
\end{tabular}
\caption{\textbf{Edit budget.} Net score of Opus on Waterbirds with ResNet-18 for different limits on the number of edit calls, and the median number of edit calls used. Each cell has 8 edit processes on the same 8 species. The two runs were made separately, and each includes the default budget of 8 as its reference.}
\label{tab:budget}
\end{table}

\Cref{tab:budget} gives the numbers behind the edit-budget figure (\cref{fig:budget}), including a budget of 2 edit calls that the figure omits. The budgets were tested in two separate runs, each with its own reference of 8 edit calls, and every comparison in the main text is made within a run. The probe and read budgets were kept at 20 and 30. With a budget of 8 or 12, Opus uses a median of 5 to 6 edit calls, so the default budget of 8 rarely binds.

\section{Further Edit Analysis}
\label{sec:supp-edits}

This section complements \cref{sec:editpatterns}.

\subsection{Edits on ERM and ERT networks for all species}

\begin{figure}[t]
\centering
\includegraphics[width=\linewidth]{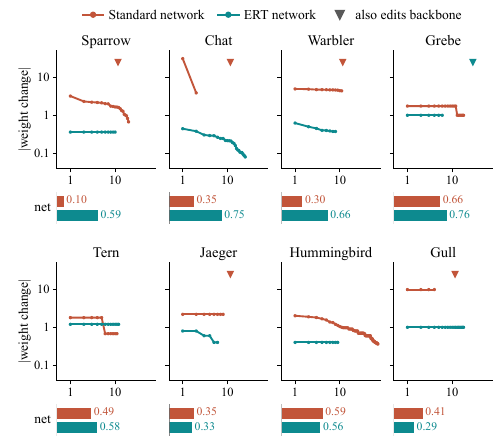}
\caption{\textbf{Opus's edits for the same request on an ERM and on an ERT network}, for all eight Waterbirds species (ResNet-18, ordered by the gain from ERT). Curves: size of the change of each edited head unit, $|\Delta(w_{\text{target}}-w_{\text{source}})|$, from largest to smallest (log axes; horizontal axis: rank of the unit). Triangles: the edit also shifts channel thresholds in the backbone. Bars: held-out net score.}
\label{fig:edit-pairs}
\end{figure}

\Cref{fig:edit-pairs} compares Opus's edits for the same request on an ERM and on an ERT network, for all eight Waterbirds species of the main grid (ResNet-18, training seed 0). None is selected. The change of the head is smaller on the ERT network for all eight species, and the held-out net score is higher on the ERT network for five of them. Opus also shifts channel thresholds in the backbone in five of the edits on ERM networks and in one of those on ERT networks.

\subsection{Head changes of all main-table edits}
\label{sec:edit-heatmaps}

\begin{figure*}[t]
\centering
\includegraphics[width=\linewidth]{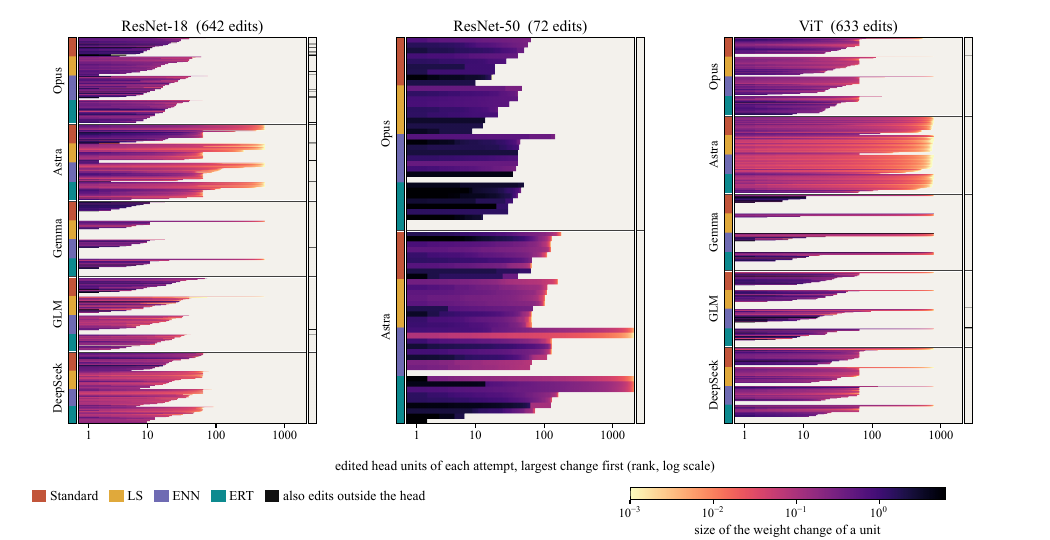}
\caption{\textbf{Head changes of all main-table edits, by backbone.} One row per edit process, grouped by editor and, within an editor, by the training method of the edited network (left strip). Horizontal axis: the edited head units, from the largest to the smallest change (rank, log scale, the same in all panels; the head has 512 units in ResNet-18, 2{,}048 in ResNet-50 and 768 in ViT-B/16). Color: size of the change of the unit (log scale). Rows without color changed no head weight. Right strip: the edit also changed a tensor outside the head.}
\label{fig:heat-backbone}
\end{figure*}

\begin{figure*}[t]
\centering
\includegraphics[width=\linewidth]{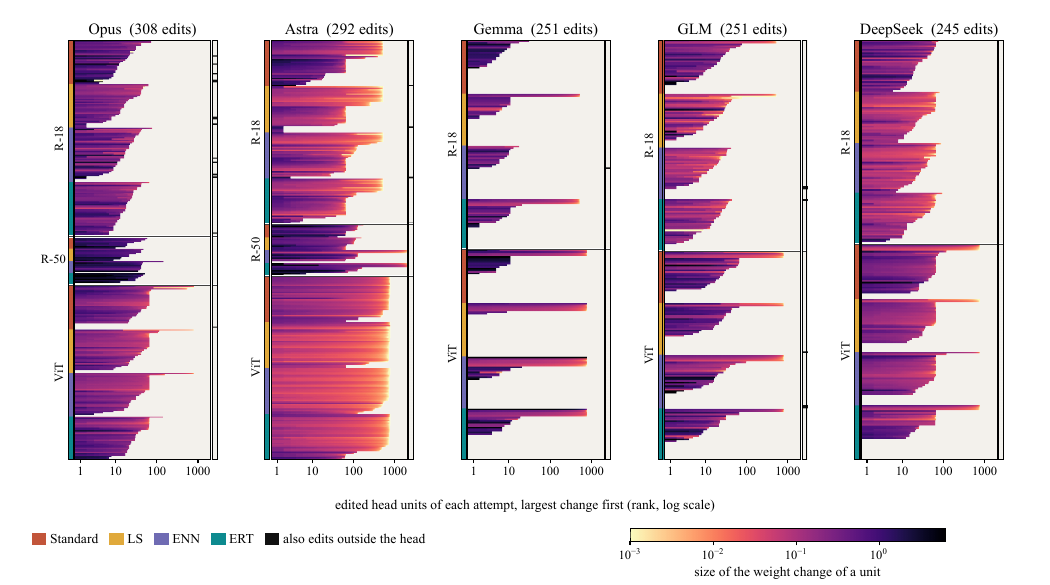}
\caption{\textbf{Head changes of all main-table edits, by editor.} The rows of \cref{fig:heat-backbone}, grouped by editor and, within an editor, by backbone (R-18, R-50: ResNet-18 and ResNet-50) and training method. Axes, colors and strips as in \cref{fig:heat-backbone}.}
\label{fig:heat-editor}
\end{figure*}

\begin{table*}[t]
\centering
\footnotesize
\setlength{\tabcolsep}{4.5pt}
\inputtab tables_bench/edit_heatmap_stats.tex
\caption{\textbf{Size of the head change over all main-table edits.} \emph{Edits}: edit processes behind the main tables with a stored edit. \emph{Head}: edit processes that changed a head weight; the statistics are taken over these. \emph{Edited units}: number of head units with a changed weight. \emph{Change of a unit}: $\ell_2$ norm over the classes of the unit's column of $\Delta W$, over all edited units of all edit processes. \emph{Change of the head}: $\|\Delta W\|_F$ per edit process. \emph{Outside}: edit processes that also changed a tensor outside the head. The same data as \cref{fig:heat-backbone,fig:heat-editor}.}
\label{tab:heat-stats}
\end{table*}

\Cref{fig:edit-pairs} covers 16 edit processes of one editor. \Cref{fig:heat-backbone,fig:heat-editor} show the same quantity for every edit process behind \cref{tab:main} and \cref{tab:others} whose edit we stored: 1{,}347 edit processes of the five editors, on 646 relabeling requests, 386 confusion fixes and 315 subgroup fixes (21 further edit processes ended without a submission). None is selected.

\paragraph{Reading the figures.} Each row is one edit process. The networks differ between edit processes, so head units cannot be aligned by their index. Instead, each row lists the units that the edit changed, from the largest to the smallest change, and the color gives the size of the change, measured as the $\ell_2$ norm over the classes of the unit's column of $\Delta W$. A row therefore ends at the number of edited units, and a row without color changed no head weight. The two figures show the same rows, grouped by backbone and by editor.

\paragraph{Editors differ more than training methods.} Opus changes few units (median 25) by a comparatively large amount. Astra changes many units by small amounts (on average 206 of 512 units on ResNet-18 and 638 of 768 on ViT-B/16) and rewrites the whole head in 62 of 292 edit processes (37 of them on ResNet-18). DeepSeek (median 45 units) and GLM (median 20) lie in between; GLM changes no head weight in 57 of 251 edit processes. Gemma leaves the weights untouched in 135 of 251 edit processes, which agrees with its low net score, and otherwise either changes a handful of units or the whole head (25 edit processes). Edits outside the head are rare for every editor: 14 of Opus's 308 edit processes also change a tensor in the backbone (13 of them on ResNet-18), 6 of GLM's, 4 of Astra's and 1 of Gemma's. Within an editor, the rows of the four training methods look alike. The training method does not change which kind of edit an editor writes, in line with \cref{sec:editpatterns}.

\paragraph{Summary statistics.} \Cref{tab:heat-stats} gives the minimum, mean and maximum of the number of edited units, of the change of a single unit and of the change of the whole head ($\|\Delta W\|_F$). The minimum of the change of a unit is bounded by the threshold of $10^{-7}$ above which we count a weight as changed, and reflects whole-head operations that leave most units almost unchanged. The means are dominated by a few very large edits, in particular of Gemma on ViT-B/16 (largest head change 5{,}693), so the lower block should not be read as a comparison of the training methods.

\subsection{Edit similarity and score}

How similar an Opus edit is to a gradient step does not tell which method wins. Over the 75 paired ENN and ERT edit processes behind the main table, no measure of this similarity is positively related to the advantage of ENN (rank correlations $-0.29$ to $+0.20$, none positive with $p<0.05$). Features of the edit also barely predict its score. Only on CIFAR-100 relabeling with ResNet-18 does the net score rise with the similarity to all three closed-form reference edits (rank correlation 0.62 to 0.65, 32 edit processes); elsewhere the correlations are at most 0.65 in absolute value, with mixed signs.

\section{Scale, Compute and Release}
\label{sec:supp-scale}

\begin{table*}[t]
\centering
\small
\setlength{\tabcolsep}{6pt}
\begin{tabular}{@{}l rrrrr r rr@{}}
\toprule
& & \multicolumn{4}{c}{Tool calls} & Changed weights & \multicolumn{2}{c}{Editing time} \\
\cmidrule(lr){3-6}\cmidrule(lr){8-9}
Editor & Edit processes & all & probes & edits & weight reads & (median) & total (h) & median (min) \\
\midrule
GPT-6 Astra & 401 & 11{,}782 & 7{,}038 & 1{,}845 & 926 & 252 & 27.2 & 3.1 \\
Claude Opus 5.5 & 1{,}098 & 25{,}354 & 13{,}882 & 5{,}646 & 609 & 27 & 65.4 & 2.7 \\
GLM-5.3-Flash & 401 & 14{,}027 & 6{,}752 & 2{,}073 & 3{,}166 & 24 & 26.4 & 2.7 \\
DeepSeek-V4.1-Flash & 315 & 10{,}831 & 5{,}789 & 1{,}749 & 1{,}211 & 44 & 104.4 & 17.1 \\
Gemma-4-26B & 448 & 14{,}638 & 6{,}996 & 2{,}770 & 1{,}425 & 10 & 11.0 & 0.7 \\
\midrule
All editors & 2{,}663 & 76{,}632 & 40{,}457 & 14{,}083 & 7{,}337 & 32 & 234.5 & 2.8 \\
\bottomrule
\end{tabular}
\caption{\textbf{Scale of the study.} Scored edit processes per editor with their logged tool calls: all calls, calls to the three probe tools, applied edit calls and calls to \tool{read\_weights}. \emph{Changed weights}: median number of weight entries that differ from the original network, over the edit processes that changed at least one. Editing time runs from the first tool call of an edit process to its submission.}
\label{tab:scale}
\end{table*}

\paragraph{Scale.} \Cref{tab:scale} counts every scored edit process of the study as of 18 November 2026: the main grid, the ablations and pilots of the main text, and earlier grids that were run while the protocol was developed. The median edit process takes about three minutes for Astra and Opus, with a median of 252 and 27 changed weights.

\paragraph{Compute.} The edit processes were made on 109 networks. Together with hyperparameter tuning, ablations and training seeds, we trained about 1{,}300 networks. Training them, including the evaluations during training, took about 385 GPU-hours (summed from the training logs), about 320 on NVIDIA RTX PRO 6000 Blackwell GPUs and about 65 on NVIDIA H100 and A100 GPUs. Scoring the edited networks adds only a few hours. The three open-weight editors consumed about 750 million prompt tokens.

\paragraph{Release.} We will release the tasks (support sets and objectives), the full agent transcripts, every tool call, the submitted weight changes and all scores of the edit processes in \cref{tab:scale}, together with the code of the WeightLab server and of the training methods.

\end{document}

%% file: bench_preamble.tex
\usepackage[T1]{fontenc}
\usepackage{graphicx}
\usepackage{amsmath,amssymb}
\usepackage{booktabs}
\usepackage{microtype}
\usepackage{multirow}
\usepackage{colortbl}
\usepackage{tikz}
\usepackage{wrapfig}
\usetikzlibrary{arrows.meta,positioning}
\definecolor{bestbg}{rgb}{0.84,0.93,0.93}

\newcommand{\editorlogo}[1]{\IfFileExists{fig_bench/logo_#1.png}{\raisebox{-0.18em}{\includegraphics[height=0.95em]{fig_bench/logo_#1.png}}}{}}
\newcommand{\cistack}[2]{\begin{tabular}[c]{@{}c@{}}#1\\[-3.5pt]{\fontsize{5.5}{6.5}\selectfont\color{black!55}[#2]}\end{tabular}}
\newcommand{\metricci}[3]{\cistack{#1}{#2,\,#3}}
\newcommand{\bestmetricci}[3]{\cistack{\textbf{#1}}{#2,\,#3}}
\newcommand{\secondmetricci}[3]{\cistack{\underline{#1}}{#2,\,#3}}
\definecolor{deltapos}{rgb}{0.00,0.45,0.25}
\definecolor{deltaneg}{rgb}{0.75,0.10,0.10}
\newcommand{\posdelta}[1]{\textcolor{deltapos}{#1}}
\newcommand{\negdelta}[1]{\textcolor{deltaneg}{#1}}
\newcommand{\sigposdelta}[1]{\textcolor{deltapos}{\boldmath\textbf{#1}}}

\newcommand{\neudelta}[1]{\textcolor{black!55}{#1}}

%% file: tables_bench/table1.tex
\begin{tabular*}{\textwidth}{@{\extracolsep{\fill}}c@{\hspace{3pt}}l*{10}{c}@{}}
\toprule
& & \multicolumn{4}{c}{\textbf{Relabel}} & \multicolumn{3}{c}{\textbf{Confusion fix}} & \multicolumn{2}{c}{\textbf{Subgroup fix}} & \\
\cmidrule(lr){3-6}\cmidrule(lr){7-9}\cmidrule(lr){10-11}
 &
  & \textbf{WB}
  & \textbf{WB}
  & \textbf{C100}
  & \textbf{C100}
  & \textbf{C100}
  & \textbf{C100}
  & \textbf{E30}
  & \textbf{PACS}
  & \textbf{PACS}
  & \\[-0.2em]
& \textbf{Method}
  & \textbf{R18}
  & \textbf{ViT}
  & \textbf{R18}
  & \textbf{ViT}
  & \textbf{R18}
  & \textbf{ViT}
  & \textbf{R50}
  & \textbf{R18}
  & \textbf{ViT}
  & \textbf{Mean} \\
\midrule
\multirow{6}{*}{\rotatebox[origin=c]{90}{\footnotesize\editorlogo{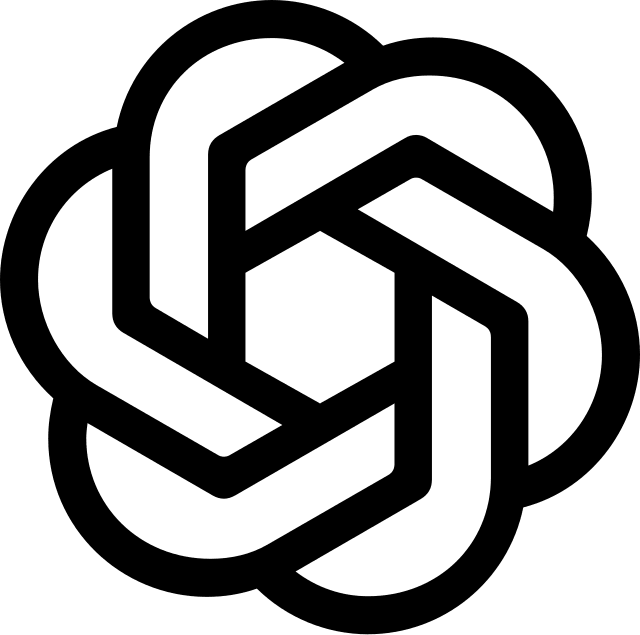}\,\textit{GPT-6 Astra}}}
& ERM \textit{(standard)}
& \metricci{.40}{.32}{.48}
& \metricci{.50}{.37}{.64}
& \metricci{.64}{.52}{.74}
& \metricci{.77}{.68}{.85}
& \metricci{.60}{.50}{.72}
& \metricci{.67}{.49}{.80}
& \metricci{.46}{.37}{.55}
& \secondmetricci{.77}{.66}{.86}
& \bestmetricci{.82}{.78}{.87}
& \metricci{.63}{.59}{.66} \\

& LS
& \metricci{.42}{.34}{.50}
& \secondmetricci{.57}{.44}{.70}
& \secondmetricci{.68}{.50}{.80}
& \metricci{.80}{.71}{.89}
& \metricci{.50}{.43}{.58}
& \secondmetricci{.74}{.61}{.86}
& \metricci{.53}{.43}{.62}
& \metricci{.71}{.57}{.83}
& \metricci{.72}{.60}{.81}
& \metricci{.63}{.59}{.67} \\

& ENN
& \secondmetricci{.47}{.42}{.52}
& \bestmetricci{.74}{.68}{.81}
& \metricci{.58}{.45}{.73}
& \bestmetricci{.85}{.76}{.92}
& \bestmetricci{.73}{.60}{.85}
& \metricci{.68}{.54}{.81}
& \bestmetricci{.81}{.72}{.89}
& \metricci{.74}{.60}{.84}
& \metricci{.78}{.74}{.82}
& \secondmetricci{.71}{.68}{.74} \\

& ERT \textit{(ours)}
& \bestmetricci{.63}{.54}{.71}
& \bestmetricci{.74}{.65}{.82}
& \bestmetricci{.71}{.53}{.84}
& \secondmetricci{.81}{.70}{.91}
& \secondmetricci{.65}{.42}{.84}
& \bestmetricci{.75}{.69}{.80}
& \secondmetricci{.79}{.71}{.87}
& \bestmetricci{.81}{.71}{.88}
& \secondmetricci{.81}{.77}{.84}
& \bestmetricci{.74}{.70}{.78} \\

\cmidrule{2-12}
& $\Delta$ vs.\ std.
& \sigposdelta{$+.23$}
& \sigposdelta{$+.24$}
& \posdelta{$+.07$}
& \posdelta{$+.05$}
& \posdelta{$+.04$}
& \posdelta{$+.08$}
& \sigposdelta{$+.33$}
& \posdelta{$+.03$}
& \negdelta{$-.02$}
& \sigposdelta{$+.12$} \\[-0.1em]

& $\Delta$ vs.\ best
& \sigposdelta{$+.16$}
& \neudelta{.00}
& \posdelta{$+.03$}
& \negdelta{$-.04$}
& \negdelta{$-.08$}
& \posdelta{$+.01$}
& \negdelta{$-.02$}
& \posdelta{$+.03$}
& \negdelta{$-.02$}
& \posdelta{$+.03$} \\
\midrule
\multirow{6}{*}{\rotatebox[origin=c]{90}{\footnotesize\editorlogo{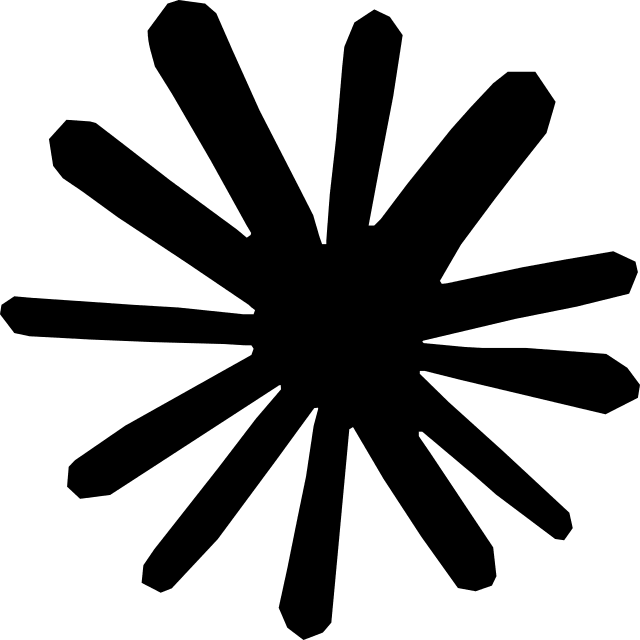}\,\textit{Claude Opus 5.5}}}
& ERM \textit{(standard)}
& \metricci{.41}{.29}{.52}
& \metricci{.33}{.21}{.46}
& \secondmetricci{.68}{.57}{.77}
& \metricci{.81}{.69}{.89}
& \metricci{.59}{.51}{.67}
& \metricci{.61}{.49}{.75}
& \metricci{.44}{.32}{.56}
& \secondmetricci{.72}{.59}{.83}
& \bestmetricci{.81}{.77}{.84}
& \metricci{.60}{.56}{.64} \\

& LS
& \metricci{.48}{.34}{.60}
& \metricci{.62}{.43}{.77}
& \bestmetricci{.80}{.77}{.85}
& \metricci{.81}{.59}{.94}
& \metricci{.48}{.36}{.60}
& \metricci{.64}{.58}{.70}
& \metricci{.51}{.39}{.62}
& \bestmetricci{.74}{.62}{.84}
& \metricci{.77}{.71}{.83}
& \metricci{.65}{.61}{.69} \\

& ENN
& \secondmetricci{.52}{.41}{.63}
& \bestmetricci{.79}{.71}{.86}
& \metricci{.66}{.59}{.73}
& \bestmetricci{.91}{.90}{.93}
& \secondmetricci{.60}{.47}{.72}
& \secondmetricci{.71}{.59}{.84}
& \secondmetricci{.60}{.50}{.71}
& \secondmetricci{.72}{.57}{.84}
& \metricci{.65}{.45}{.79}
& \secondmetricci{.68}{.65}{.72} \\

& ERT \textit{(ours)}
& \bestmetricci{.63}{.54}{.71}
& \secondmetricci{.76}{.68}{.83}
& \metricci{.67}{.52}{.80}
& \secondmetricci{.86}{.81}{.90}
& \bestmetricci{.61}{.46}{.73}
& \bestmetricci{.74}{.65}{.82}
& \bestmetricci{.74}{.63}{.83}
& \metricci{.70}{.55}{.83}
& \secondmetricci{.80}{.77}{.84}
& \bestmetricci{.72}{.69}{.76} \\

\cmidrule{2-12}
& $\Delta$ vs.\ std.
& \posdelta{$+.16$}
& \sigposdelta{$+.44$}
& \neudelta{.00}
& \posdelta{$+.05$}
& \posdelta{$+.02$}
& \sigposdelta{$+.12$}
& \sigposdelta{$+.30$}
& \negdelta{$-.02$}
& \negdelta{$-.01$}
& \sigposdelta{$+.12$} \\[-0.1em]

& $\Delta$ vs.\ best
& \sigposdelta{$+.10$}
& \negdelta{$-.02$}
& \negdelta{$-.13$}
& \negdelta{$-.05$}
& \posdelta{$+.01$}
& \posdelta{$+.03$}
& \posdelta{$+.14$}
& \negdelta{$-.04$}
& \negdelta{$-.01$}
& \posdelta{$+.04$} \\
\bottomrule
\end{tabular*}